\documentclass[10pt,twocolumn,letterpaper]{article}

\usepackage{authblk}
\usepackage{cvpr}
\usepackage[export]{adjustbox}
\usepackage{amsfonts}       %
\usepackage{amsmath}        %
\usepackage{amssymb}
\usepackage{appendix}
\usepackage{array}
\usepackage{booktabs}       %
\usepackage{epsfig}
\usepackage[T1]{fontenc}    %
\usepackage{graphicx}
\usepackage[utf8]{inputenc} %
\usepackage{microtype}      %
\usepackage{nicefrac}       %
\usepackage{tabularx}
\usepackage{times}
\usepackage{url}            %
\usepackage[usenames,dvipsnames]{xcolor}
\usepackage{xspace}
\usepackage{dblfloatfix}

\usepackage{enumitem}
\setlist{noitemsep,topsep=0pt,parsep=0pt,partopsep=0pt,leftmargin=*}

\usepackage{multirow}
\usepackage{colortbl}
\usepackage{subcaption}
\usepackage{siunitx}

\usepackage[font=small]{caption}

\makeatletter
\@namedef{ver@everyshi.sty}{} %
\makeatother

\usepackage{tikz}
\usepackage{pgfplots}
\usetikzlibrary{spy,calc}
\newif\ifblackandwhitecycle
\gdef\patternnumber{0}

\pgfkeys{/tikz/.cd,
    zoombox paths/.style={
        draw=orange,
        very thick
    },
    black and white/.is choice,
    black and white/.default=static,
    black and white/static/.style={
        draw=white,
        zoombox paths/.append style={
            draw=white,
            postaction={
                draw=black,
                loosely dashed
            }
        }
    },
    black and white/static/.code={
        \gdef\patternnumber{1}
    },
    black and white/cycle/.code={
        \blackandwhitecycletrue
        \gdef\patternnumber{1}
    },
    black and white pattern/.is choice,
    black and white pattern/0/.style={},
    black and white pattern/1/.style={
            draw=white,
            postaction={
                draw=black,
                dash pattern=on 2pt off 2pt
            }
    },
    black and white pattern/2/.style={
            draw=white,
            postaction={
                draw=black,
                dash pattern=on 4pt off 4pt
            }
    },
    black and white pattern/3/.style={
            draw=white,
            postaction={
                draw=black,
                dash pattern=on 4pt off 4pt on 1pt off 4pt
            }
    },
    black and white pattern/4/.style={
            draw=white,
            postaction={
                draw=black,
                dash pattern=on 4pt off 2pt on 2 pt off 2pt on 2 pt off 2pt
            }
    },
    zoomboxarray inner gap/.initial=5pt,
    zoomboxarray columns/.initial=2,
    zoomboxarray rows/.initial=1,
    zoomboxarray heightmultiplier/.initial=0.5,
    subfigurename/.initial={},
    figurename/.initial={zoombox},
    zoomboxarray/.style={
        execute at begin picture={
            \begin{scope}[
                spy using outlines={%
                    zoombox paths,
                    width=\imagewidth / \pgfkeysvalueof{/tikz/zoomboxarray columns} - (\pgfkeysvalueof{/tikz/zoomboxarray columns} - 1) / \pgfkeysvalueof{/tikz/zoomboxarray columns} * \pgfkeysvalueof{/tikz/zoomboxarray inner gap} -\pgflinewidth,
                    height=\pgfkeysvalueof{/tikz/zoomboxarray heightmultiplier} * (\imageheight / \pgfkeysvalueof{/tikz/zoomboxarray rows} - (\pgfkeysvalueof{/tikz/zoomboxarray rows} - 1) / \pgfkeysvalueof{/tikz/zoomboxarray rows} * \pgfkeysvalueof{/tikz/zoomboxarray inner gap}-\pgflinewidth),
                    magnification=3,
                    every spy on node/.style={
                        zoombox paths
                    },
                    every spy in node/.style={
                        zoombox paths
                    }
                }
            ]
        },
        execute at end picture={
            \end{scope}
     \gdef\patternnumber{0}
        },
        spymargin/.initial=0.5em,
        zoomboxes xshift/.initial=1,
        zoomboxes right/.code=\pgfkeys{/tikz/zoomboxes xshift=1},
        zoomboxes left/.code=\pgfkeys{/tikz/zoomboxes xshift=-1},
        zoomboxes yshift/.initial=0,
        zoomboxes above/.code={
            \pgfkeys{/tikz/zoomboxes yshift=1},
            \pgfkeys{/tikz/zoomboxes xshift=0}
        },
        zoomboxes below/.code={
            \pgfkeys{/tikz/zoomboxes yshift=-1},
            \pgfkeys{/tikz/zoomboxes xshift=0}
        },
        caption margin/.initial=0ex, %
    },
    adjust caption spacing/.code={},
    image container/.style={
        inner sep=0pt,
        at=(image.north),
        anchor=north,
        adjust caption spacing
    },
    zoomboxes container/.style={
        inner sep=0pt,
        at=(image.north),
        anchor=north,
        name=zoomboxes container,
        xshift=\pgfkeysvalueof{/tikz/zoomboxes xshift}*(\imagewidth+\pgfkeysvalueof{/tikz/spymargin}),
        yshift=\pgfkeysvalueof{/tikz/zoomboxes yshift}*(\imageheight+\pgfkeysvalueof{/tikz/spymargin}+\pgfkeysvalueof{/tikz/caption margin}),
        adjust caption spacing
    },
    calculate dimensions/.code={
        \pgfpointdiff{\pgfpointanchor{image}{south west} }{\pgfpointanchor{image}{north east} }
        \pgfgetlastxy{\imagewidth}{\imageheight}
        \global\let\imagewidth=\imagewidth
        \global\let\imageheight=\imageheight
        \gdef\columncount{1}
        \gdef\rowcount{1}
        
    },
    image node/.style={
        inner sep=0pt,
        name=image,
        anchor=south west,
        append after command={
            [calculate dimensions]
            node [image container,subfigurename=\pgfkeysvalueof{/tikz/figurename}-image] {\phantomimage}
            node [zoomboxes container,subfigurename=\pgfkeysvalueof{/tikz/figurename}-zoom] {\phantomimage}
        }
    },
    color code/.style={
        zoombox paths/.append style={draw=#1}
    },
    connect zoomboxes/.style={
    spy connection path={\draw[draw=none,zoombox paths] (tikzspyonnode) -- (tikzspyinnode);}
    },
    help grid code/.code={
        \begin{scope}[
                x={(image.south east)},
                y={(image.north west)},
                font=\footnotesize,
                help lines,
                overlay
            ]
            \foreach \x in {0,1,...,9} {
                \draw(\x/10,0) -- (\x/10,1);
                \node [anchor=north] at (\x/10,0) {0.\x};
            }
            \foreach \y in {0,1,...,9} {
                \draw(0,\y/10) -- (1,\y/10);                        \node [anchor=east] at (0,\y/10) {0.\y};
            }
        \end{scope}
    },
    help grid/.style={
        append after command={
            [help grid code]
        }
    },
}

\newcommand\phantomimage{%
    \phantom{%
        \rule{\imagewidth}{\imageheight}%
    }%
}
\newcommand\zoombox[2][]{
    \begin{scope}[zoombox paths]
        \pgfmathsetmacro\xpos{
            (\columncount-1)*(\imagewidth / \pgfkeysvalueof{/tikz/zoomboxarray columns} + \pgfkeysvalueof{/tikz/zoomboxarray inner gap} / \pgfkeysvalueof{/tikz/zoomboxarray columns} ) + \pgflinewidth
        }
        \pgfmathsetmacro\ypos{
            (\rowcount-1) * (\imageheight / \pgfkeysvalueof{/tikz/zoomboxarray rows} + \pgfkeysvalueof{/tikz/zoomboxarray inner gap} / \pgfkeysvalueof{/tikz/zoomboxarray rows} ) + 0.5*\pgflinewidth
        }
        \edef\dospy{\noexpand\spy [
            #1,
            zoombox paths/.append style={
                black and white pattern=\patternnumber
            },
            every spy on node/.append style={#1},
            x=\imagewidth,
            y=\imageheight
        ] on (#2) in node [anchor=north west] at ($(zoomboxes container.north west)+(\xpos pt,-\ypos pt)$);}
        \dospy
        \pgfmathtruncatemacro\pgfmathresult{ifthenelse(\columncount==\pgfkeysvalueof{/tikz/zoomboxarray columns},\rowcount+1,\rowcount)}
        \global\let\rowcount=\pgfmathresult
        \pgfmathtruncatemacro\pgfmathresult{ifthenelse(\columncount==\pgfkeysvalueof{/tikz/zoomboxarray columns},1,\columncount+1)}
        \global\let\columncount=\pgfmathresult
        \ifblackandwhitecycle
            \pgfmathtruncatemacro{\newpatternnumber}{\patternnumber+1}
            \global\edef\patternnumber{\newpatternnumber}
        \fi
    \end{scope}
}

\usepackage{xr-hyper}

\usepackage[pagebackref=true,breaklinks=true,letterpaper=true,colorlinks,bookmarks=false]{hyperref} %

\cvprfinalcopy %

\def\cvprPaperID{3702} %

\pgfplotsset{compat=1.15}
\begin{document}

\let\originalleft\left
\let\originalright\right
\def\left#1{\mathopen{}\originalleft#1}
\def\right#1{\originalright#1\mathclose{}}

\definecolor{yellow}{rgb}{1,1, 0.6}
\definecolor{lightorange}{rgb}{1, 0.8, 0.6}
\definecolor{lightred}{rgb}{0.9, 0.5, 0.5}
\definecolor{olive}{rgb}{0.0, 0.7, 0.0}
\definecolor{alexey}{rgb}{0.8, 0.0, 0.8}

\definecolor{niceblue}{rgb}{0., 0.635294118, 1.0}
\definecolor{nicered}{rgb}{1.0, 0.392156863, 0.319672131}
\definecolor{azure}{rgb}{0.0, 0.3, 0.6}
\definecolor{bestazure}{HTML}{3867d6}
\definecolor{secondazure}{HTML}{1474b8}

\newcommand\BEST[1]{\textcolor{bestazure}{\textbf{#1}}}
\newcommand\SECOND[1]{\textcolor{secondazure}{#1}}

\newcommand\CVPRTODO[1]{\textcolor{azure}{[CVPR-TODO: #1]}}

\newcommand\TODO[1]{\textcolor{blue}{[TODO: #1]}}
\newcommand\mehdi[1]{\textcolor{olive}{[MS: #1]}}
\newcommand\daniel[1]{\textcolor{purple}{[DD: #1]}}
\newcommand\barron[1]{\textcolor{lightred}{[JB: #1]}}
\newcommand\noha[1]{\textcolor{brown}{[NR: #1]}}
\newcommand\rmbrualla[1]{\textcolor{orange}{[RMB: #1]}}
\newcommand\alexey[1]{\textcolor{alexey}{[AD: #1]}}

\makeatletter
\newcommand{\minisection}{%
  \@startsection{paragraph}{4}%
  {\z@}{1.0ex \@plus 1ex \@minus .2ex}{-1em}%
  {\normalfont \normalsize \bfseries}%
}
\makeatother
\def \NERFW {NeRF-W}
\def \NERFU {NeRF-U}
\def \NERFG {NeRF-A}
\def \NERF  {NeRF}
\def \Inthewild  {In-the-wild}
\def \inthewild  {in-the-wild}
\def \ie {\emph{i.e}\onedot}
\def \eg {\emph{e.g}\onedot}

\newcommand{\normsq}[1]{\big\lVert#1\big\rVert^2_2}
\newcommand\embed[0]{\boldsymbol{\ell}}
\newcommand{\ray}{\mathbf{r}}
\newcommand{\col}{\mathbf{c}}
\newcommand{\truecol}{\mathbf{C}}
\newcommand{\Col}{\mathbf{C}}
\newcommand{\position}{\mathbf{x}}
\newcommand{\vieworigin}{\mathbf{o}}
\newcommand{\viewdir}{\mathbf{d}}
\newcommand{\latentpos}{\mathbf{z}}
\newcommand{\posenc}{\gamma_{\position}}
\newcommand{\viewenc}{\gamma_{\viewdir}}
\newcommand{\mlp}{\operatorname{MLP}}
\newcommand{\image}{\mathcal{I}}
\newcommand{\camera}{\mathbf{p}}
\newcommand{\nerfurl}{\url{https://TODO}}
\newcommand\transient[1]{#1^{(\tau)}}
\newcommand\static[1]{#1}
\newcommand\coarse[1]{#1^{c}}
\newcommand\fine[1]{#1^{f}}
\newcommand\approximate[1]{\hat{#1}}
\newcommand\decay[1]{\alpha \left( #1 \right)}
\newcommand\glo[1]{#1^{(a)}}
\newcommand\uncertainty[1]{\transient{#1}}
\newcommand{\nerfmlp}{\operatorname{MLP}_{\text{NeRF}}}
\newcommand\flickrcc[1]{\small{Photos by Flickr users #1 / \href{https://creativecommons.org/licenses/by/2.0/}{CC BY}}}
\newcommand\flickrccone[1]{\small{Photo by Flickr user #1 / \href{https://creativecommons.org/licenses/by/2.0/}{CC BY}}}
\newcommand\blendercc[1]{\small{Render by Blender Swap user #1 / \href{https://creativecommons.org/licenses/by/2.0/}{CC BY}}}
\newcommand{\renderingprocess}{\mathcal{R}}

\newcommand{\secref}[1]{Section~\ref{#1}}
\newcommand{\chapref}[1]{Chapter~\ref{#1}}
\renewcommand{\eqref}[1]{Equation~(\ref{#1})}
\newcommand{\figref}[1]{Figure~\ref{#1}}
\newcommand{\tabref}[1]{Table~\ref{#1}}
\newcommand{\algref}[1]{Algorithm~\ref{#1}}

\renewcommand{\UrlFont}{\ttfamily\small}

\title{NeRF in the Wild: Neural Radiance Fields for Unconstrained Photo Collections}

\newcommand{\authortt}[1]{{\tt\small #1}}
\newcommand\CoAuthorMark{\footnotemark[\arabic{footnote}]}
\author{Ricardo Martin-Brualla\thanks{Denotes equal contribution.}}
\author{\, Noha Radwan\protect\CoAuthorMark}
\author{\, Mehdi S. M. Sajjadi\protect\CoAuthorMark} 
\author{\,  \quad\quad\quad\quad\quad\quad\quad\quad\quad\quad\quad\quad\quad\quad\quad\quad Jonathan T. Barron}
\author{\, Alexey~Dosovitskiy}
\author{Daniel Duckworth}
\vspace{-3pt}
\affil{Google Research \authorcr 
    \authortt{\{rmbrualla, noharadwan, msajjadi, barron, adosovitskiy, duckworthd\}@google.com}}

\maketitle

\begin{abstract}
We present a learning-based method for synthesizing novel views of complex scenes using only unstructured collections of \inthewild{} photographs.
We build on Neural Radiance Fields (\NERF{}), which uses the weights of a multilayer perceptron to model the density and color of a scene as a function of 3D coordinates.
While \NERF{} works well on images of static subjects captured under controlled settings, it is incapable of modeling many ubiquitous, real-world phenomena in uncontrolled images, such as variable illumination or transient occluders.
We introduce a series of extensions to \NERF{} to address these issues, thereby enabling accurate reconstructions from unstructured image collections taken from the internet.
We apply our system, dubbed \NERFW{}, to internet photo collections of famous landmarks, and demonstrate temporally consistent novel view renderings that are significantly closer to photorealism than the prior state of the art.
\end{abstract}
\vspace{-2mm}

\section{Introduction} \label{sec:introduction}

Synthesizing novel views of a scene from a sparse set of captured images is a long-standing problem in computer vision, and a prerequisite to many AR and VR applications.
Though classic techniques have addressed this problem using structure-from-motion~\cite{hartley2003multiple} or image-based rendering~\cite{shum2008image}, this field has recently seen significant progress due to \emph{neural rendering} techniques --- learning-based modules embedded within a 3D geometric context, and trained to reconstruct observed images.
The Neural Radiance Fields (\NERF{}) approach~\cite{mildenhall2020nerf} models the radiance field and density of a scene with the weights of a neural network.
Volume rendering is then used to synthesize new views, demonstrating a heretofore unprecedented level of fidelity on a range of challenging scenes.
However, \NERF{} has only been demonstrated to work well in controlled settings: the scene is captured within a short time frame during which lighting effects remain constant, and all content in the scene is static.
As we will demonstrate, \NERF{}'s performance degrades significantly when presented with moving objects or variable illumination.
This limitation prohibits direct application of \NERF{} to large-scale \inthewild{} scenarios, where input images may be taken hours or years apart, and may contain pedestrians and vehicles moving through them.

\begin{figure}[t]
    \centering 
    \setlength{\tabcolsep}{2pt}
    \def\arraystretch{1}
    \begin{tabular}{cc}
        \includegraphics[width=0.22\columnwidth]{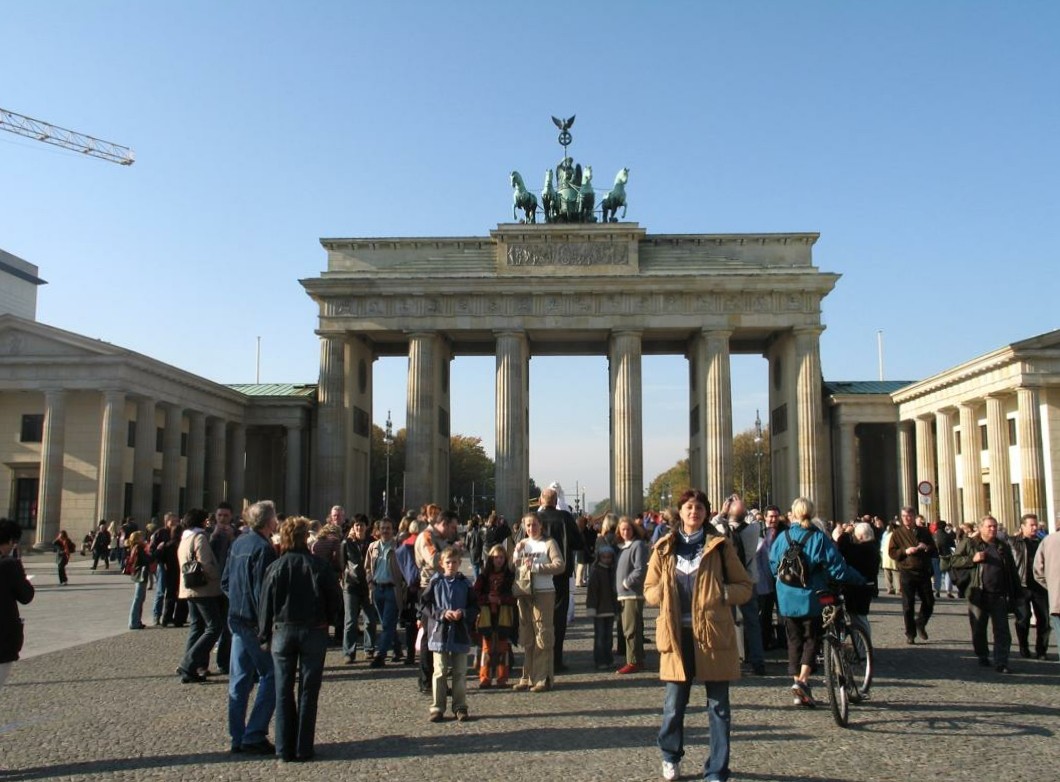} &
        \multirow{3}{*}[31pt]{\includegraphics[width=0.74\columnwidth]{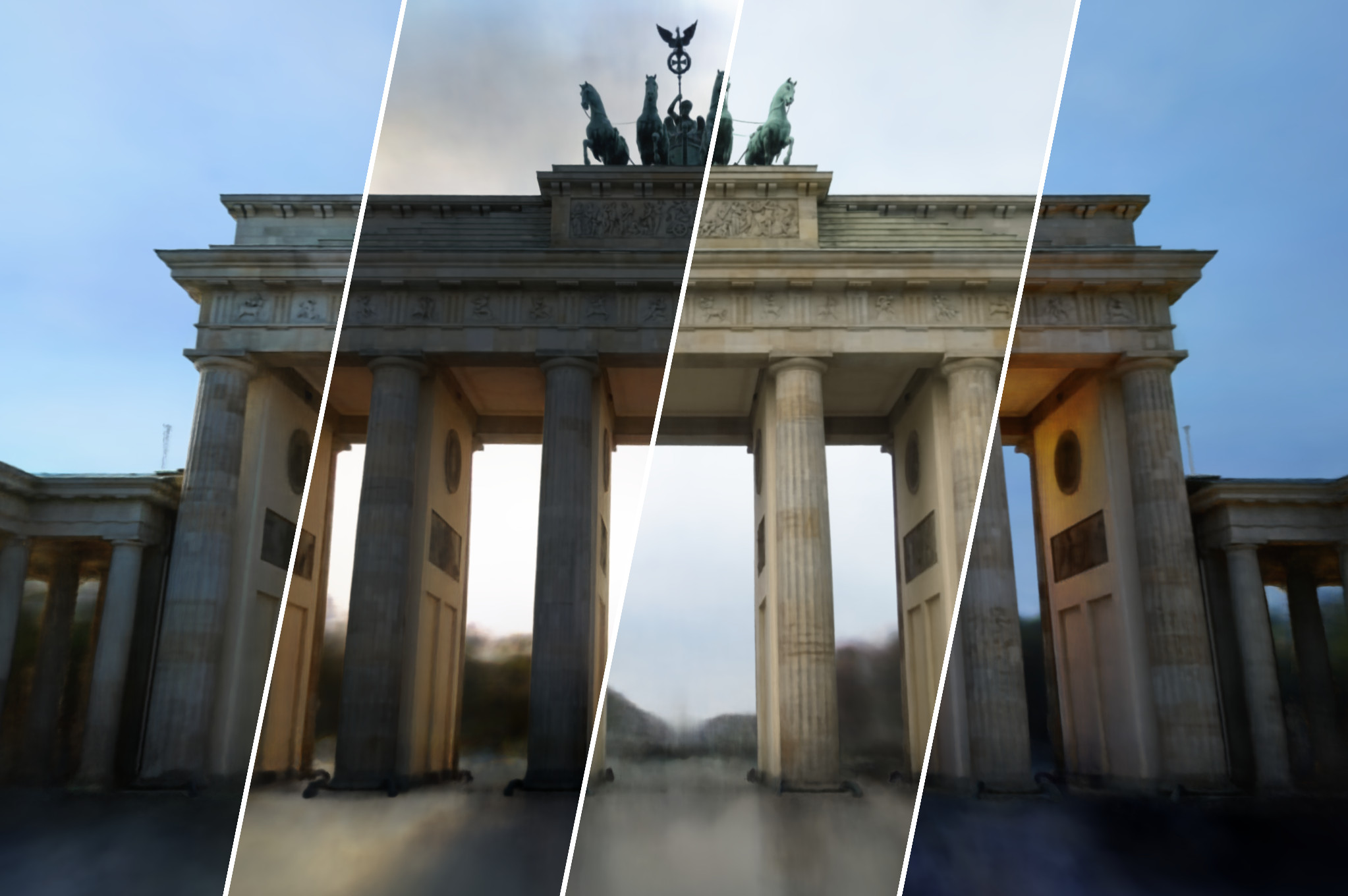}} \\
        \includegraphics[width=0.22\columnwidth]{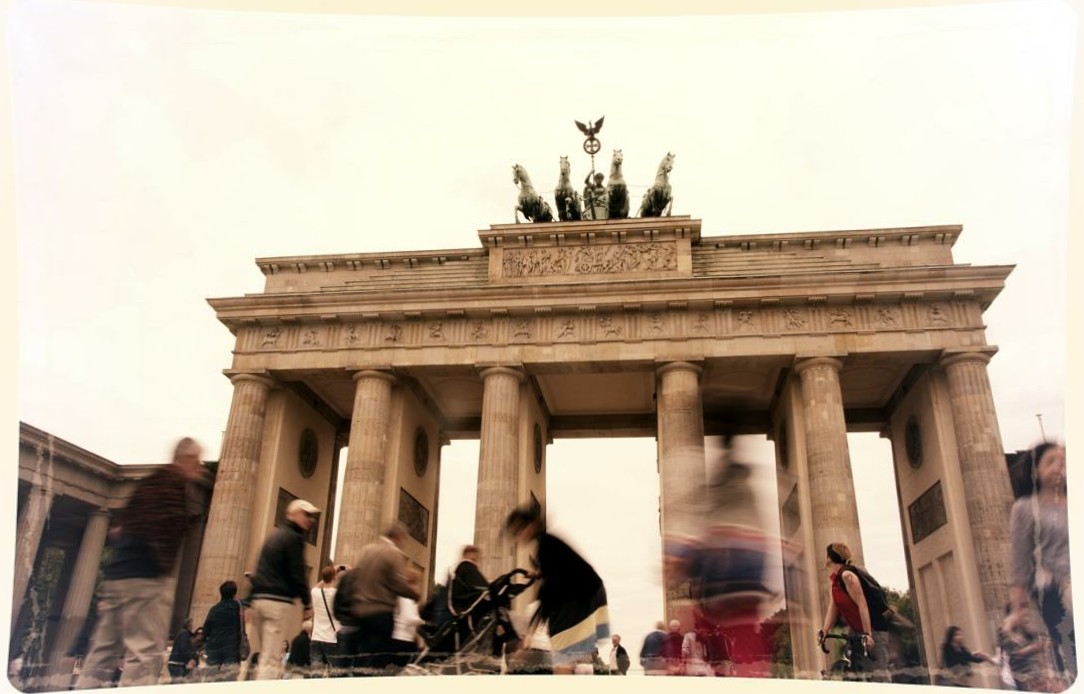} &\\
        \includegraphics[width=0.22\columnwidth]{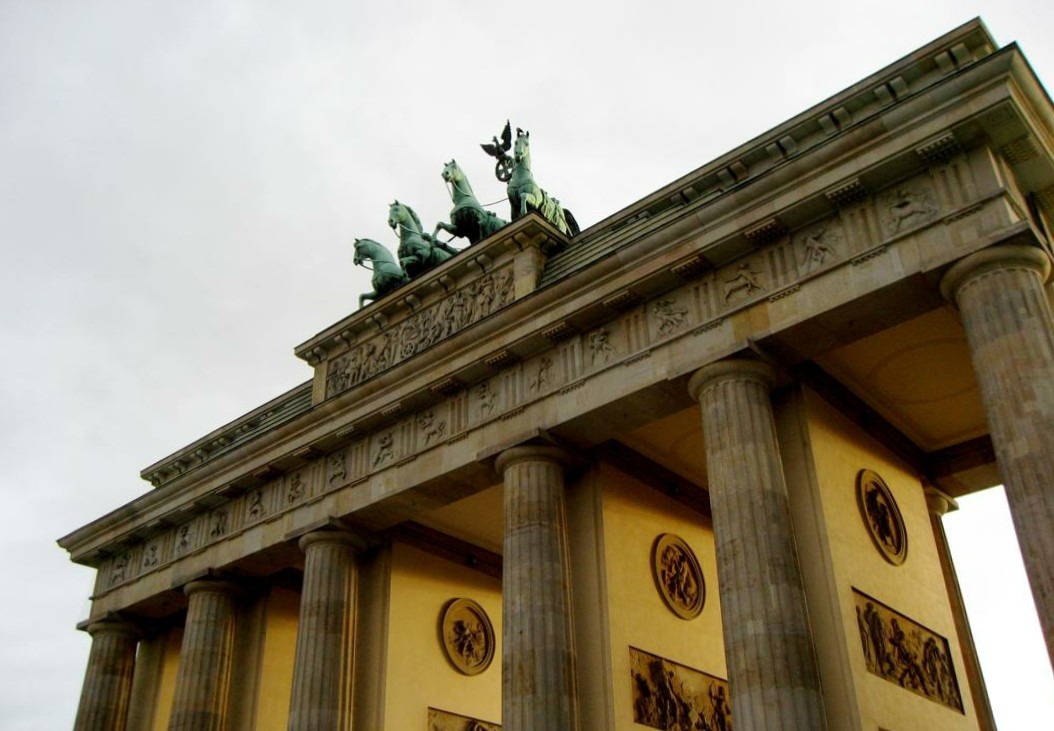} &\\
        \small (a) Photos &
        \small (b) Renderings
    \end{tabular}
    \vspace{-0.1in}
    \caption{
    Given only an internet photo collection (a), our method is able to render novel views with variable illumination (b).
    \flickrcc{dbowie78, vasnic64, punch}.
    }
    \label{fig:cover_figure}
\end{figure}

The central limitation of \NERF{} that we address here is its assumption that the world is geometrically, materially, and photometrically \emph{static} --- that the density and radiance of the world is constant. 
\NERF{} therefore requires that any two photographs taken at the same position and orientation must be identical.
This assumption is severely violated in many real-world datasets, such as large-scale internet photo collections of tourist landmarks.
Two photographers may stand in the same location and photograph the same landmark, but in the time between those two photographs the world can change significantly: cars and people may move, construction may begin or end, seasons and weather may change, the sun may move through the sky, etc.
Even two photos taken at the same time and location can exhibit considerable variation: exposure, color correction, and tone-mapping all may vary depending on the camera and post-processing.
We will demonstrate that naively applying \NERF{} to \inthewild{} photo collections results in inaccurate reconstructions that exhibit severe ghosting, oversmoothing, and further artifacts.

To handle these demanding scenarios, we present \NERFW{}, an extension of \NERF{} that relaxes its strict consistency assumptions.
First, we model per-image appearance variations such as exposure, lighting, weather, and post-processing in a learned low-dimensional latent space.
Following the framework of Generative Latent Optimization \cite{bojanowski2017optimizing}, we optimize an appearance embedding for each input image, thereby granting \NERFW{} the flexibility to explain away photometric and environmental variations between images by learning a shared appearance representation across the entire photo collection.
The learned latent space provides control of the appearance of output renderings as illustrated in Figure~\ref{fig:cover_figure}, (b).
Second, we model the scene as the union of shared and image-dependent elements, thereby enabling the unsupervised decomposition of scene content into ``static'' and ``transient'' components.
Our approach models transient elements using a secondary volumetric radiance field combined with a data-dependent uncertainty field, where the latter captures variable observation noise and further reduces the effect of transient objects on the static scene representation.
Because optimization is able to identify and discount transient image content, we can synthesize realistic renderings of novel views by rendering only the static component.

We apply \NERFW{} to several challenging \inthewild{} photo collections of cultural landmarks and show that it can produce detailed, high-fidelity renderings from novel viewpoints, surpassing the prior state of the art by a large margin on PSNR and MS-SSIM.
Unlike prior work, renderings from our model exhibit smooth appearance interpolation and temporal consistency, even for wide camera trajectories.
We find that \NERFW{} significantly improves quality over \NERF{} in the presence of appearance variation and transient occluders while achieving similar quality in controlled settings.

\newcommand{\challengespic}[1]{
    \adjincludegraphics[max width=0.33\columnwidth]{challenges/#1.jpg}
}

\begin{figure}
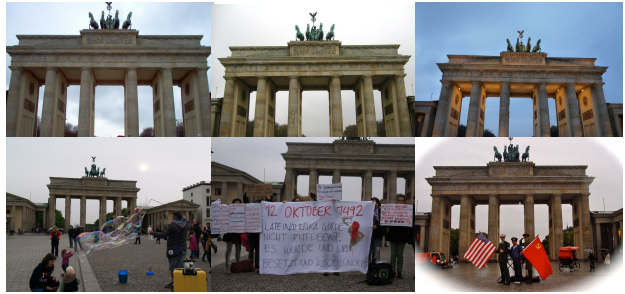

    \centering 
    \setlength{\tabcolsep}{1pt}
    \def\arraystretch{1}
    \begin{tabular}{@{}c@{\!\!}c@{\!\!}c@{}}
        \challengespic{000045} &
        \challengespic{000214} &
        \challengespic{000085}
        \\ [-4mm]
        \challengespic{000005} &
        \challengespic{000036} &
        \challengespic{000110} 
    \\
    \end{tabular} 
    \vspace{-0.1in}
    \caption{
        Example \inthewild{} photographs from the Phototourism dataset~\cite{jin2020image} used to train \NERFW{}.
        Due to variable illumination and post-processing (top), the same object's color may vary from image to image.
        \Inthewild{} photos may also contain transient occluding subjects (bottom).
        \flickrcc{paradasos, itia4u, jblesa, joshheumann, ojotes, chyauchentravelworld}.
    }
    \label{fig:challenges-examples}
\end{figure}

\section{Related Work} \label{sec:related-work}

The last decade has seen the integration of physics-based multi-view geometry techniques into deep learning-based approaches for the task of 3D scene reconstruction.
Here we review recent progress on novel view synthesis and neural rendering, and highlight the main differences between existing approaches and our proposed method.

\minisection{Novel View Synthesis:}
Constructing novel views of a scene captured by multiple images is a long standing problem in computer vision. Structure-from-Motion~\cite{hartley2003multiple} and bundle adjustment~\cite{triggs1999bundle} can be used to reconstruct a sparse point cloud representation and recover camera parameters. 
Photo Tourism~\cite{snavely2006photo} showed how these reconstruction techniques could be scaled to unconstrained photo collections and used to perform view synthesis~\cite{agarwal2011romeinaday,frahm2010cloudlessrome}.
Other approaches to view synthesis include light-field photography~\cite{levoy1996light} and image-based rendering~\cite{Buehler2001} but these generally require a dense capture of the scene.
Recent works explicitly infer the light and reflectance properties of the objects in the scene from a set of unconstrained photo collections~\cite{laffont2012coherent,shan2013visual}, 
and other utilize semantic knowledge to reconstruct transient objects~\cite{price2018augmenting}.

\minisection{Neural Rendering:}
More recently, neural rendering techniques~\cite{Tewari2020stateofartneuralrendering} have been applied to scene reconstruction. 
Several approaches employ image translation networks \cite{pix2pix2017} to re-render content more realistically using as input traditional reconstruction results~\cite{martinbrualla2018lookingood}, learned latent textures~\cite{thies2019deferred}, point clouds~\cite{aliev2019neural}, voxels~\cite{sitzmann2019deepvoxels}, or plane sweep volumes~\cite{flynn2019deepview, flynn2016deepstereo}. 
Most similar in application to our work is Neural Rerendering in the Wild (NRW)~\cite{meshry2019neural} which synthesizes realistic novel views of tourist sites from point cloud renderings by learning a neural re-rendering network conditioned on a learned latent appearance embedding module. 
Common drawbacks of these approaches are the checkerboard and temporal artifacts visible under camera motion caused by the employed 2D image translation network.
Another recent approach represents the scene as camera-centric multiplane images to reconstruct captured scenes~\cite{mildenhall2019llff, zhou2018stereo}, and internet photo collections~\cite{li2020crowdsampling}.
These methods produce photorealistic renderings of novel viewpoints but the views they can interpolate are restricted to a small volume surrounding the ground truth camera poses.
In contrast, volume rendering approaches~\cite{Lombardi_2019, mildenhall2020nerf, sitzmann2019scene} allow for accurate and consistent reconstructions even with large camera motions, as does \NERFW{}.
Neural Radiance Fields (NeRF) \cite{mildenhall2020nerf} use a multi-layer perceptron (MLP) to model a radiance field at an unprecedented level of fidelity, in part thanks to the use of positional encoding within the MLP~\cite{tancik2020fourier}. 
Our work focuses on extending NeRF to unconstrained scenarios, like internet photo collections.

\section{Background}
\label{sec:background}
\label{sec:nerf}

Our goal is to produce a system that takes as input a photo collection and then learns a 3D representation that is capable of generating the photos of that collection.
Such a scene representation should encode the 3D structure of the scene together with appearance information so as to enable the synthesis of novel, unseen views.
In the following we describe Neural Radiance Fields \cite{mildenhall2020nerf} (\NERF{}), the method for 3D scene reconstruction that \NERFW{} extends.

\begin{figure}[t!]
    \centering
    \includegraphics[width=0.9\linewidth]{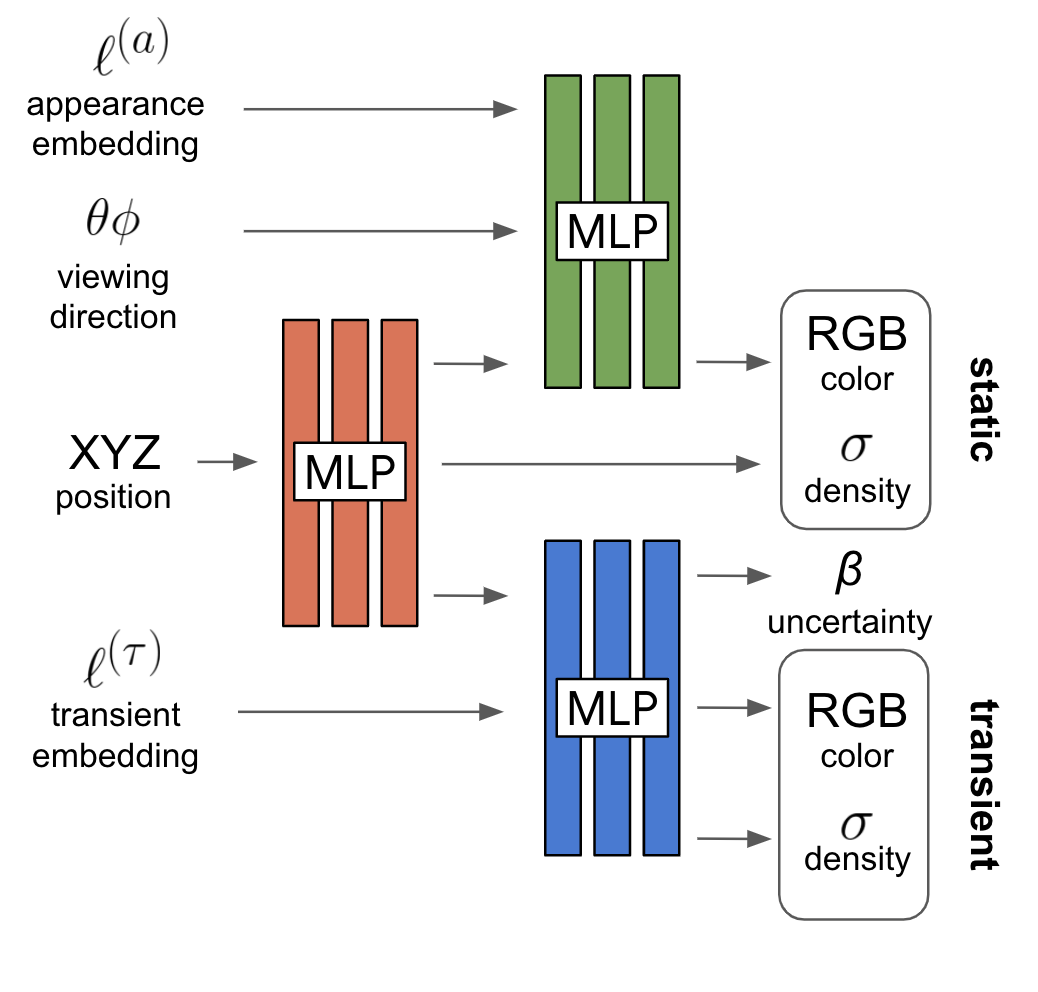}
    \caption{
        \NERFW{} model architecture.
        Given a 3D position, viewing direction, and learned appearance and transient embeddings, \NERFW{} produces static and transient colors and densities as well as a measure of uncertainty.
        Note that the static opacity is generated \emph{before} the model is conditioned on the appearance embedding, ensuring that static geometry is shared across all images. %
    }
    \label{fig:model-arch}
\end{figure}

\NERF{} represents a scene using a learned, continuous volumetric radiance field $F_{\theta}$ defined over a bounded 3D volume.
$F_{\theta}$ is modeled using a multilayer perceptron (MLP) that takes as input a 3D position $\position = (x, y, z)$ and unit-norm viewing direction $\viewdir = (d_x, d_y, d_z)$,  and produces as output a density $\sigma$ and color $\col = (r, g, b)$.
To compute the color of a single pixel, \NERF{} approximates the volume rendering integral using numerical quadrature~\cite{max1995optical}.
Let $\ray(t) = \vieworigin + t \viewdir$ be the camera ray emitted from the center of projection of a camera $\vieworigin$ through a given pixel on the image plane.
\NERF{}'s approximation of the expected color $\approximate{\Col}(\ray)$ of that pixel is:
\begin{gather}
    \approximate{\Col}(\ray) = \renderingprocess(\ray, \col , \sigma) = 
        \sum_{k=1}^{K} T(t_k) \, \decay{\sigma(t_k) \delta_{k}} \, \col(t_k) \, , \label{eqn:nerf1}\\
    \text{where} \quad T(t_k) = 
        \exp \left(-\sum_{k'=1}^{k-1} \sigma(t_{k'}) \delta_{k'} \right) \, ,\label{eqn:nerf2}
\end{gather}
where $\renderingprocess(\ray, \col , \sigma)$ is the volumetric rendering of color $\col$ with density $\sigma$, $\col(t)$ and $\sigma(t)$  are the color and density at point $\ray(t)$, $\decay{x} = 1-\exp(-x)$, and $\delta_k = t_{k+1} - t_k$ is the distance between two quadrature points. Stratified sampling is used to select quadrature points $\{ t_k \}_{k=1}^{K}$ between $t_n$ and $t_f$, the near and far planes of the camera.

\newcommand{\nerfwbreakdownpic}[1]{
    \includegraphics[width=0.195\textwidth]{nerfw_breakdown/brandenburg_v2/#1.jpg}
}

\begin{figure*}[h!]
    \centering 
    \setlength{\tabcolsep}{-0.5pt}
    \def\arraystretch{1}
    \begin{tabular}{ccccc}
        \nerfwbreakdownpic{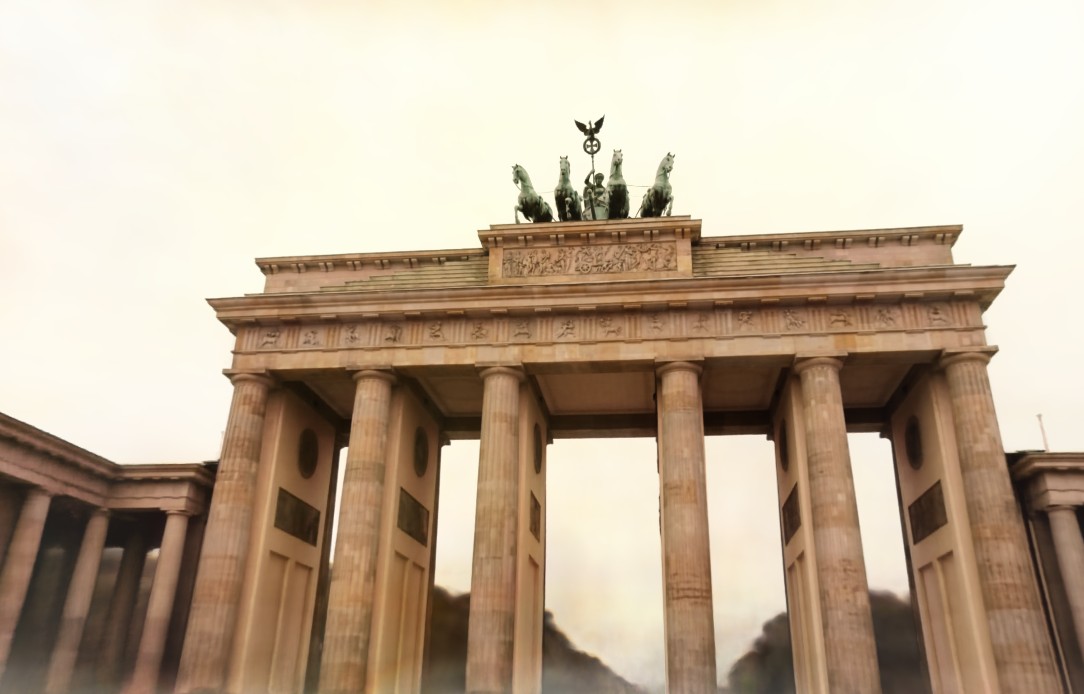} &
        \nerfwbreakdownpic{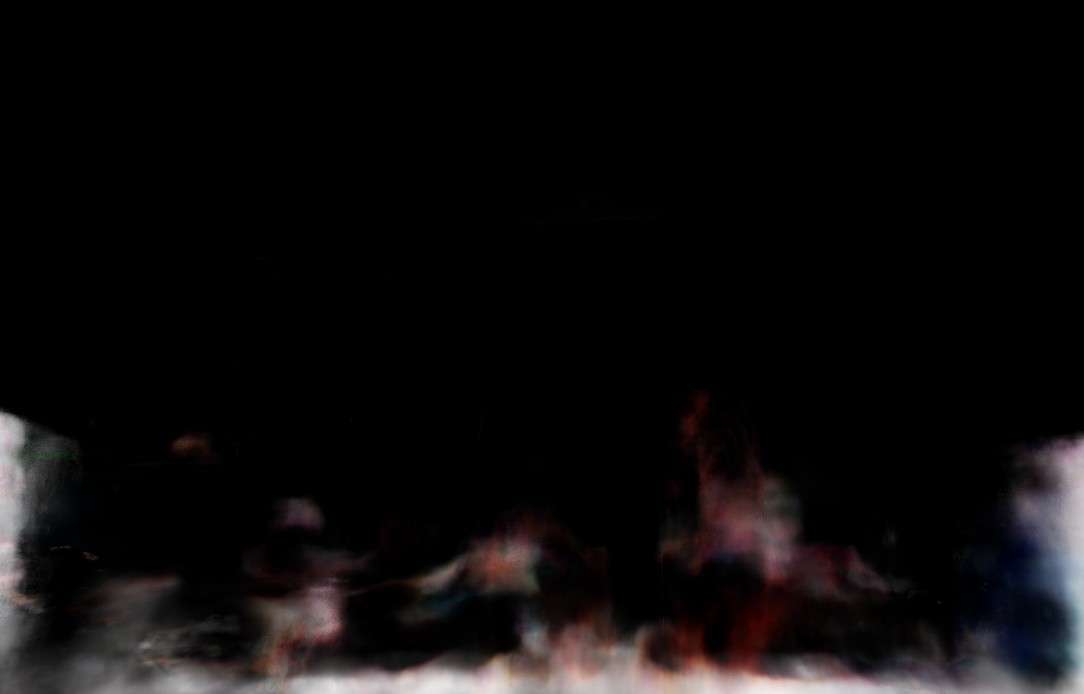} &
        \nerfwbreakdownpic{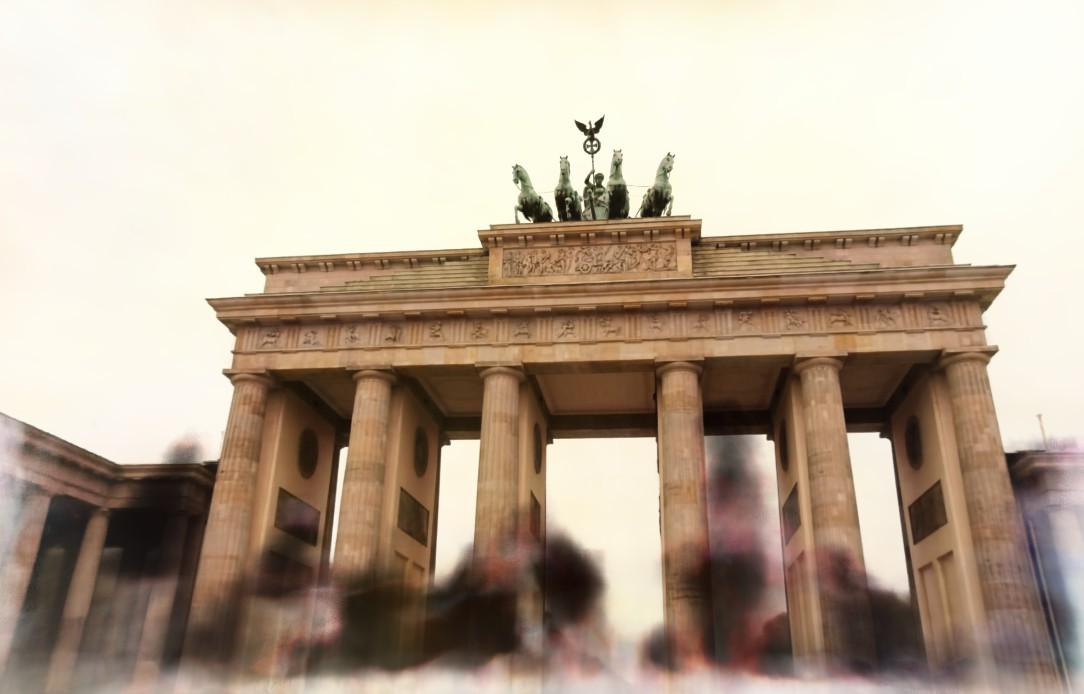} &
        \nerfwbreakdownpic{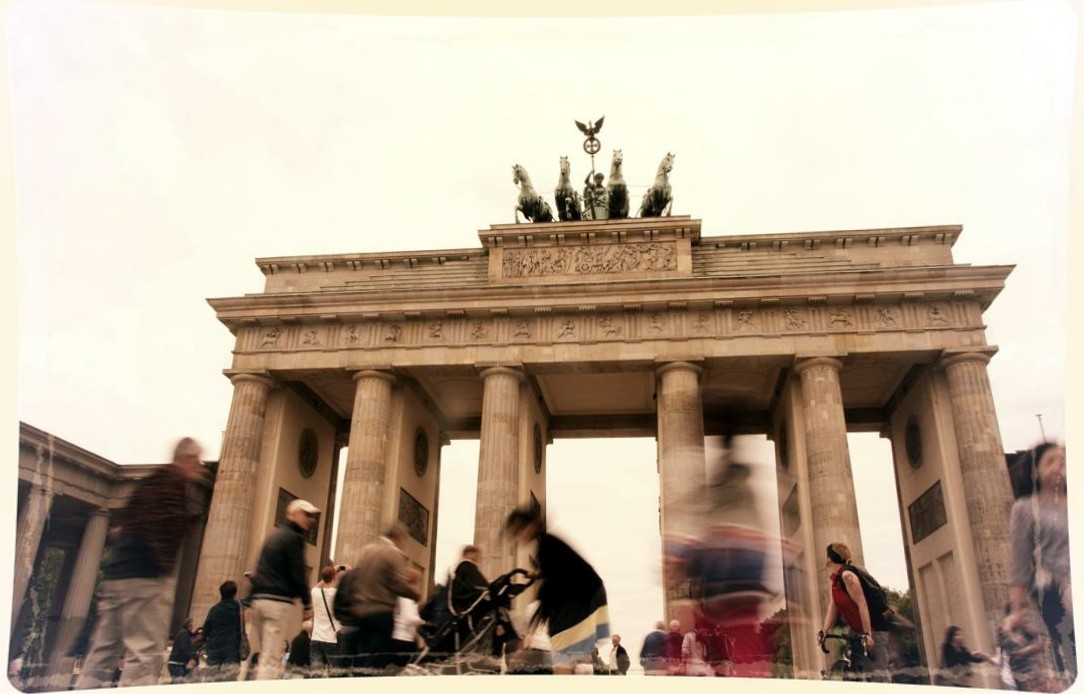} &
        \nerfwbreakdownpic{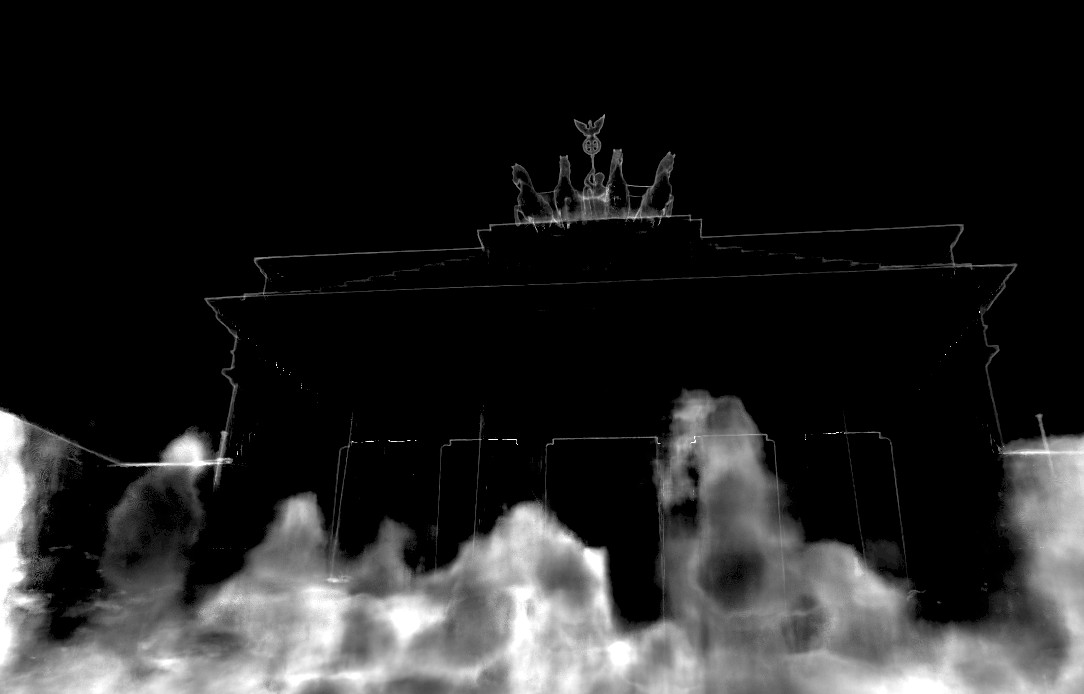}
        \\
        \small (a) Static &
        \small (b) Transient &
        \small (c) Composite &
        \small (d) Image &
        \small (e) Uncertainty
        \\
    \end{tabular} 
    \vspace{-0.1in}
    \caption{
        \NERFW{} separately renders the static (a) and transient (b) elements of the scene, and then composites them (c). 
        Training minimizes the difference between the composite and the true image (d) weighted by uncertainty (e), which is simultaneously optimized to identify and discount anomalous image regions. 
        \flickrccone{vasnic64}.
    }
    \label{fig:nerfw-breakdown}
\end{figure*}

\NERF{} represents the volumetric density $\sigma(t)$ and color $\col(t)$ using ReLU MLPs of the following form:
\begin{align}
    \left[ \sigma(t), \latentpos(t) \right] &=
        \mlp_{\theta_1}\left( \posenc\left(\ray(t)\right) \right)\, , \\
    \col(t) &=
        \mlp_{\theta_2}\left( \latentpos(t), \viewenc(\viewdir) \right)\, , \label{eqn:nerf-mlp2}
\end{align}
with parameters $\theta = [\theta_1, \theta_2]$ and fixed encoding functions $\posenc$ (for position) and $\viewenc$ (for viewing direction).
The final activations in generating $\sigma(t)$ and $\col(t)$ are a ReLU and a sigmoid respectively, as density must be non-negative and color must be in $[0, 1]$.
Unlike \cite{mildenhall2020nerf}, we describe the neural network as two MLPs where the latter depends on one output of the former, $\latentpos(t)$, to highlight the fact that volume density $\sigma(t)$ is independent of viewing direction $\viewdir$.

To fit parameters $\theta$, \NERF{} minimizes the sum of squared reconstruction errors with respect to an RGB image collection $\{ \image_i \}_{i=1}^{N}$, $\image_i \in [0, 1]^{H \times W \times 3}$.
Each image $\image_i$ is paired with its corresponding intrinsic and extrinsic camera parameters which can be estimated using structure-from-motion~\cite{schoenberger2016sfm}.
We precompute the set of camera rays $\{ \ray_{ij} \}_{j=1}^{H \times W \times 3}$ corresponding to pixel $j$ from image $i$ with each ray passing through the 3D location $\vieworigin_{i}$ with direction $\viewdir_{ij}$, where $\ray_{ij}(t) = \vieworigin_{i} + t \, \viewdir_{ij}$. 

To improve sample efficiency, \NERF{} simultaneously optimizes two MLPs: one coarse and one fine, where the density predicted by the coarse model determines the sampling of quadrature points for the fine model.
The parameters of both models are optimized by minimizing the following loss:
\begin{equation}
    \sum_{ij}
        \normsq{\truecol(\ray_{ij}) - \coarse{\approximate{\Col}}(\ray_{ij})} 
    + \normsq{\truecol(\ray_{ij}) - \fine{\approximate{\Col}}(\ray_{ij})}\, ,
\end{equation}
where $\truecol(\ray_{ij})$ is the observed color of ray $j$ in image $\image_{i}$,
and $\coarse{\approximate{\Col}}$ and $\fine{\approximate{\Col}}$ are the coarse and fine models respectively.

\section{NeRF in the Wild} \label{sec:method}

We now present \NERFW{}, a system for reconstructing 3D scenes from \inthewild{} photo collections.
We build on \NERF{}~\cite{mildenhall2020nerf} and introduce two enhancements explicitly designed to handle the challenges of unconstrained imagery.

Similar to \NERF{}, we learn a volumetric density representation $F_{\theta}$ from an unstructured photo collection $\{ \image_i \}_{i=1}^{N}$ for which camera parameters are known.
\NERF{} assumes consistency in its input views: that a point in 3D space observed from the same position and viewing direction in two different images has the same intensity. 
But this assumption is violated by internet photos (such as those shown in Figure~\ref{fig:challenges-examples}) due to two distinct phenomena:

\noindent{\textbf{1) Photometric variation:}} In outdoor photography, time of day and atmospheric conditions directly impact the illumination (and consequently, the emitted radiance) of objects in the scene. 
This issue is exacerbated by photographic imaging pipelines, as variation in auto-exposure settings, white balance, and tone-mapping across photographs may result in additional photometric inconsistencies~\cite{BrooksCVPR2019}. 

\noindent{\textbf{2) Transient objects:}} Real-world landmarks are rarely captured in isolation, without moving objects or occluders around them. 
Tourist photos of landmarks are particularly challenging, as they often contain posing human subjects and other pedestrians. 

We propose two model components to address these issues.
In~\secref{sec:glo} we extend \NERF{} to allow for image-dependent appearance and illumination variations such that photometric discrepancies between images can be modeled explicitly.
In~\secref{sec:transient-objects} we further extend this model by allowing transient objects to be jointly estimated and disentangled from a static representation of the 3D world.
Figure~\ref{fig:model-arch} shows an overview of the proposed model architecture.

\subsection{Latent Appearance Modeling} \label{sec:glo}

To adapt \NERF{} to variable lighting and photometric post-processing,
we adopt the approach of Generative Latent Optimization (GLO)~\cite{bojanowski2017optimizing} in which each image $\image_i$ is assigned a corresponding real-valued appearance embedding vector $\glo{\embed}_i $ of length $\glo{n}$.
We replace the image-independent radiance $\col(t)$ in \eqref{eqn:nerf1} with an image-dependent radiance $\col_{i}(t)$, which also introduces a dependency on image index $i$ to the approximated pixel color $\approximate{\Col}_i$:
\begin{gather}
    \approximate{\Col}_i(\ray) = \renderingprocess(\ray, \col_{i}, \sigma) , \label{eqn:nerf-glo1} \\
    \col_{i}(t) = \mlp_{\theta_2}\left( \latentpos(t), \viewenc(\viewdir), \glo{\embed}_{i} \right) \, .
    \label{eqn:nerf-glo-mlp}
\end{gather}
The $\{ \glo{\embed}_{i} \}_{i=1}^{N}$ embeddings are optimized alongside $\theta$.

Using these appearance embeddings as input to only the branch of the network that emits color grants our model the freedom to vary the emitted radiance of the scene in a particular image while still guaranteeing that the 3D geometry (predicted earlier by $\mlp_{\theta_1}$) is static and shared across all images.
By setting $\glo{n}$ to a small value, we encourage optimization to identify a continuous space in which illumination conditions can be embedded, thereby enabling smooth interpolations between conditions as demonstrated in Figure~\ref{fig:glo-interp}.

\subsection{Transient Objects} \label{sec:transient-objects}

We address transient phenomena using two distinct design decisions:
First, we designate the color-emitting MLP (\eqref{eqn:nerf-mlp2}) used in \NERF{} as the ``static'' head of our model, and we add an additional ``transient'' head that emits its own color \emph{and density}, where that density is allowed to vary across training images.
This enables \NERFW{} to reconstruct images containing occluders without introducing artifacts into the static scene representation.
Second, instead of assuming that all observed pixel colors are equally reliable, we allow our transient head to emit a field of \emph{uncertainty} (much like our existing fields of color and density), which allows our model to adapt its reconstruction loss to ignore unreliable pixels and 3D locations that are likely to contain occluders.
We model each pixel's color as an isotropic normal distribution whose likelihood we will maximize, and we ``render'' the variance of that distribution using the same volume rendering approach used by \NERF{}.
These two model components allow \NERFW{} to disentangle static and transient phenomena without explicit supervision.

To construct our transient head, we build on the volume rendering formulation of~\eqref{eqn:nerf-glo1} and augment the static density $\sigma(t)$ and radiance $\col_{i}(t)$ with transient counterparts $\transient{\sigma}_{i}(t)$ and $\transient{\col}_{i}(t)$,
\begin{gather}
    \resizebox{0.88\linewidth}{!}{%
    $
    \displaystyle\!\!\!\!\!\!\!\approximate{\Col}_{i}(\ray)\!=\!\sum_{k=1}^{K} T_{i}(t_k) \left(\! \decay{\static{\sigma}(t_k) \delta_{k}} \static{\col}_{i}(t_k) + \decay{\!\transient{\sigma}_{i}(t_k) \delta_{k}\!} \transient{\col}_{i}(t_k) \!\right) \, , \label{eqn:nerf-uncertainty1}
    $} \\
    \resizebox{0.88\linewidth}{!}{%
    $
    \displaystyle \text{where} \,\, T_{i}(t_k) = 
        \exp \left(-\sum_{k'=1}^{k-1} \left( \sigma(t_{k'}) + \transient{\sigma}_{i}(t_{k'}) \right) \delta_{k'} \right) \, . \label{eqn:nerf-uncertainty2} 
    $}
\end{gather}
The expected color of $\ray(t)$ then becomes the alpha composite of both the static and the transient components.

We employ the Bayesian learning framework of Kendall et al.~\cite{KendallNIPS2017} to model the uncertainty of the observed color.
We assume that observed pixel intensities are inherently noisy (aleatoric) and further that this noise is input-dependent (heteroscedastic).
We model the observed color $\truecol_{i}(\ray)$ with an isotropic normal distribution with image- and ray-dependent variance $\beta_{i}(\ray)^2$ and mean $\approximate{\Col}_{i}(\ray)$.
Variance $\beta_{i}(\ray)$ is ``rendered'' analogously to color via alpha-compositing according to the transient density $\transient{\sigma}_{i}(t)$:
\begin{equation}
     \hat{\beta}_{i}(\ray) = \renderingprocess(\ray, \beta_{i}, \transient{\sigma}_{i}) .
\end{equation}
To allow the transient component of the scene to vary across images, we assign each training image $\image_i$ a second embedding $\uncertainty{\embed}_{i} \in \mathbb{R}^{\uncertainty{n}}$ that is given as input to the transient MLP,
\begin{gather}
    \left[ \transient{\sigma}_{i}(t), \transient{\col}_{i}(t), \tilde{\beta}_{i}(t) \right] =
        \mlp_{\theta_3}\left( \latentpos(t), \uncertainty{\embed}_{i} \right)\, , \label{eqn:nerf-uncertainty-mlp3} \\
    \beta_{i}(t) = \beta_{\mathrm{min}} + \log \left(1 + \exp \left( \tilde{\beta}_{i}(t) \right) \right) \, ,
\end{gather}
ReLU and sigmoid activations are used for $\transient{\sigma}_{i}(t)$ and $\transient{\col}_{i}(t)$, and a softplus is used as the activation for $\beta_{i}(t)$ (shifted by $\beta_{\min} > 0$, a hyperparameter that ensures a minimum importance is assigned to each ray).
See Figure~\ref{fig:model-arch} for an illustration of our complete model architecture.

The loss for ray $\ray$ in image $i$ with true color $\truecol_{i}(\ray)$ is
\begin{equation}
    \resizebox{0.88\linewidth}{!}{%
    $
    \displaystyle \!\!\!\!\! L_{i}(\ray) = \frac{\normsq{ \truecol_{i}(\ray) - \approximate{\Col}_{i}(\ray)}}{2 \beta_{i}(\ray)^2} + \frac{\log \beta_{i}(\ray)^2}{2}  + \frac{\lambda_{u}}{K} \sum_{k=1}^{K} \transient{\sigma}_{i}(t_{k})
    \,.
    $}
    \label{eqn:nerf-uncertainty-loss}
\end{equation}
The first two terms are the (shifted) negative log likelihood of $\truecol_{i}(\ray)$ according to a normal distribution with mean $\approximate{\Col}_{i}(\ray)$ and variance $\beta_{i}(\ray)^2$.
Larger values of $\beta_{i}(\ray)$ attenuate the importance assigned to a pixel, under the assumption that it belongs to some transient object.
The first term is balanced by the second, which corresponds to the log-partition function of the normal distribution and precludes the trivial minimum at $\beta_{i}(\ray) = \infty$.
The third term is an $L_1$ regularizer with a multiplier $\lambda_{u}$ on (non-negative) transient density $\transient{\sigma}_{i}(t)$, and this discourages the model from using transient density to explain away static phenomena.

At test time we omit the transient and uncertainty fields, and render only $\static{\sigma}(t)$ and $\static{\col}(t)$.
See Figure~\ref{fig:nerfw-breakdown} for an illustration of static, transient, and uncertainty components.

\subsection{Optimization} \label{sec:optimization}

Like \NERF{}, we simultaneously optimize two copies of $F_{\theta}$:
A fine model that uses the model and losses described above, and a coarse model that uses only the latent appearance modeling component.
Alongside parameters $\theta$ we optimize per-image appearance embeddings $\{ \glo{\embed}_{i} \}_{i=1}^{N}$ and transient embeddings $\{ \uncertainty{\embed}_{i} \}_{i=1}^{N}$. 
\NERFW{}'s loss function is then,
\begin{align}
    \sum_{ij}
        L_{i}(\ray_{ij})
        + \frac{1}{2}\normsq{\truecol(\ray_{ij}) - \coarse{\approximate{\Col}_i}(\ray_{ij})}
        \, ,
\end{align}
$\lambda_u$, $\beta_{\mathrm{min}}$, and embedding dimensionalities $\glo{n}$ and $\uncertainty{n}$ form the set of additional hyperparameters for \NERFW{}.

As optimization only produces appearance embeddings $\{ \glo{\embed}_{i} \}$ for images in the training set, the embeddings of test-set images are unspecified.
For test-set visualizations, we choose $\glo{\embed}$ to best fit a target image (\eg Figure~\ref{fig:glo-interp}) or set it to an arbitrary value.
\newcommand\validoptpic[1]{
    \raisebox{-0.5\height}{
    \begin{tikzpicture}[zoomboxarray, zoomboxes below, connect zoomboxes]
        \node[image node]{\includegraphics[width=0.31\columnwidth]{valid_opt/#1.jpg}};
        \zoombox[magnification=4,color code=nicered]{0.1,0.38}
        \zoombox[magnification=6,color code=niceblue]{0.53,0.75}
    \end{tikzpicture}}
}

\begin{figure}
    \centering 
    \setlength{\tabcolsep}{-0.5pt}
    \def\arraystretch{1}
    \begin{tabular}{@{}ccc@{}}
        \validoptpic{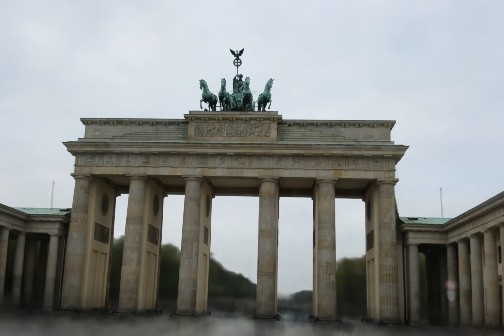} &
        \validoptpic{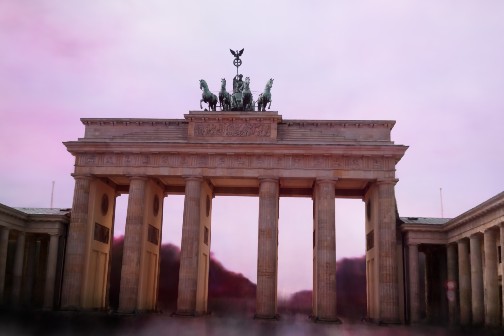} &
        \validoptpic{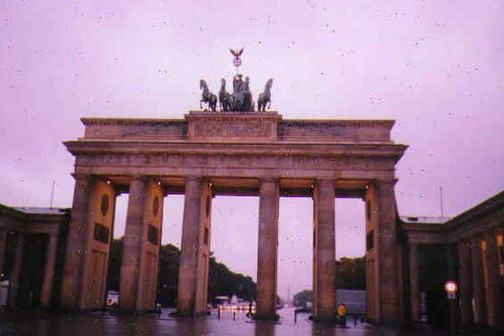}
        \\[-6mm]
        \small (a) \NERFW{} w/o opt. & 
        \small (b) \NERFW{} &
        \small (c) Reference
    \end{tabular} 
    \caption{
    Because optimization only yields appearance embeddings $\glo{\embed}$ for images in the training set, when evaluating error metrics on test-set images we optimize $\glo{\embed}$ to match the appearance of the true image using only the left half of each image.
    Error metrics are evaluated on only the right half of each image, so as to avoid information leakage.
    \flickrccone{eadaoinflynn}.
    }
    \label{fig:valid-opt}
\end{figure}

\section{Experiments} \label{sec:experiments}

Here we provide an evaluation of \NERFW{} on unconstrained (\eg ``\inthewild{}'') internet photo collections of cultural landmarks. 
We select six landmarks from the Phototourism dataset~\cite{jin2020image}.
Inspired by prior work~\cite{meshry2019neural}, we reconstruct the \emph{Trevi Fountain} and \emph{Sacre Coeur} as well as four novel scenes, the \emph{Brandenburg Gate}, \emph{Taj Mahal}, \emph{Prague Old Town Square}, and \emph{Hagia Sophia}.
Empirical performance for these scenes can be found in Table~\ref{tab:phototourism_half}, but we urge the reader to visually inspect the video results in the supplement.

\minisection{Baselines:}
We evaluate our proposed method against Neural Rerendering in the Wild (NRW)~\cite{meshry2019neural}, \NERF{}~\cite{mildenhall2020nerf}, and two ablations of \NERFW{}: \NERFG{} (appearance), wherein the ``transient'' head is eliminated; and \NERFU{} (uncertainty), wherein the appearance embedding $\glo{\embed}_{i}$ is eliminated.
\NERFW{} is the composition of \NERFG{} and \NERFU{}.
While other recent work such as \cite{li2020crowdsampling} is employed on a similar domain, we restrict baselines to those capable of extrapolating significantly beyond the views represented in the dataset.

\minisection{Optimization:}
Building on \NERF{}\footnote{\url{https://github.com/bmild/nerf}}, we implement all experiments in TensorFlow~2 using Keras. 
For each scene, we use COLMAP~\cite{schoenberger2016sfm} with two radial and two tangential distortion parameters enabled to estimate each image's camera parameters.
As in \NERF{}, for each scene we train a model initialized to random weights.
We optimize all \NERF{} variants for \num[group-separator={,}]{300000} steps with a batch size of $2048$ on 8 GPUs using Adam~\cite{kingma2015adam} (with hyperparameters $\beta_1 = 0.9$, $\beta_2=0.999$, $\epsilon=10^{-7}$), which takes approximately 2 days.
Hyperparameters shared by all \NERF{} variants are chosen to maximize PSNR on the Brandenburg Gate dataset and are fixed to those values in all other scenes.
Additional hyperparameters for variants of \NERFW{} are chosen via grid search to maximize PSNR on a held-out validation set on the Brandenburg Gate scene and are fixed to those values for all other scenes. See the supplement for additional details on hyperparameters.

\minisection{Evaluation:}
We evaluate on the task of novel view synthesis: given a held-out image with accompanying camera parameters, we render an image from the same pose and compare it to the ground truth.
As measuring perceptual image similarity is challenging~\cite{pappas2000perceptual, thung2009survey, wang2002universal, zhang2018unreasonable}, we present rendered images for visual inspection and report quantitative results based on PSNR, MS-SSIM~\cite{wang2003multiscale}, and LPIPS~\cite{zhang2018unreasonable}.
Because optimization only produces appearance embeddings for training-set images, when computing error metrics on test-set images we optimize an appearance embedding $\glo{\embed}$ on the left half of each image and report metrics on the right half (\figref{fig:valid-opt}).
See the supplement for additional discussion of error metrics.

\begin{table*}[b!]
\centering
    \setlength{\tabcolsep}{3pt}
    \begin{sc}
    \resizebox{\textwidth}{!}{%
    \begin{tabular}{l|c@{\,}c@{\,}c|c@{\,}c@{\,}c|c@{\,}c@{\,}c|c@{\,}c@{\,}c|c@{\,}c@{\,}c|c@{\,}c@{\,}c}
    &\multicolumn{3}{c|}{\footnotesize Brandenburg Gate}&\multicolumn{3}{c|}{\footnotesize Sacre Coeur}&\multicolumn{3}{c|}{\footnotesize Trevi Fountain}&\multicolumn{3}{c|}{\footnotesize Taj Mahal}&\multicolumn{3}{c|}{\footnotesize Prague}&\multicolumn{3}{c}{\footnotesize Hagia Sophia}\\
	&{\scriptsize PSNR}&{\scriptsize MS-SSIM}&{\scriptsize LPIPS}&{\scriptsize PSNR}&{\scriptsize MS-SSIM}&{\scriptsize LPIPS}&{\scriptsize PSNR}&{\scriptsize MS-SSIM}&{\scriptsize LPIPS}&{\scriptsize PSNR}&{\scriptsize MS-SSIM}&{\scriptsize LPIPS}&{\scriptsize PSNR}&{\scriptsize MS-SSIM}&{\scriptsize LPIPS}&{\scriptsize PSNR}&{\scriptsize MS-SSIM}&{\scriptsize LPIPS}\\
	\hline
    	NRW~\cite{meshry2019neural} & 23.85        & 0.914        & 0.141        & 19.39        & 0.797        & 0.229        & 20.56        & 0.811        & 0.242        & 21.24        & 0.844        & \BEST{0.201} & 19.89        & 0.803        & \BEST{0.216} & 20.75        & 0.796        & \BEST{0.231} \\
    	\NERF{}                     & 21.05        & 0.895        & 0.208        & 17.12        & 0.781        & 0.278        & 17.46        & 0.778        & 0.334        & 15.77        & 0.697        & 0.427        & 15.67        & 0.747        & 0.362        & 16.04        & 0.749        & 0.338 \\
    	\NERFG{}                    & 27.96        & 0.941        & 0.145        & 24.43        & 0.923        & 0.174        & 26.24        & 0.924        & 0.211        & 25.99        & 0.893        & 0.225        & 22.52        & 0.870        & 0.244        & 21.83        & 0.820        & 0.276 \\
    	\NERFU{}                    & 19.49        & 0.921        & 0.174        & 15.99        & 0.826        & 0.223        & 15.03        & 0.795        & 0.277        & 10.23        & 0.778        & 0.373        & 15.03        & 0.787        & 0.315        & 13.74        & 0.706        & 0.376 \\
    	\NERFW{}                    & \BEST{29.08} & \BEST{0.962} & \BEST{0.110} & \BEST{25.34} & \BEST{0.939} & \BEST{0.151} & \BEST{26.58} & \BEST{0.934} & \BEST{0.189} & \BEST{26.36} & \BEST{0.904} & 0.207        & \BEST{22.81} & \BEST{0.879} & 0.227        & \BEST{22.23} & \BEST{0.849} & 0.250 \\
	\end{tabular}%
	}
	\end{sc}
	\vspace{-0.1in}
    \caption{
        Quantitative results on the Phototourism dataset~\cite{jin2020image} for NRW~\cite{meshry2019neural}, \NERF{}~\cite{mildenhall2020nerf}, and two ablations of the proposed model.
        Best results are \BEST{highlighted}.
        \NERFW{} outperforms the previous state of the art across all datasets on PSNR and MS-SSIM and achieves competitive results in LPIPS. Note that LPIPS generally favours methods such as NRW trained with an adversarial or perceptual loss and it is less sensitive to typical GAN artifacts, see Figures~\ref{fig:phototourism_qualitative} and \ref{fig:phototourism_qualitative_2} (supplementary).
    }
    \label{tab:phototourism_half}
\end{table*}

\newcommand\depthmapbrandenburgpic[1]{
    \includegraphics[width=0.32\linewidth]{depth/brandenburg_#1.jpg}
}

\newcommand\depthmapsacrepic[1]{
    \includegraphics[width=0.32\linewidth]{depth/sacre_#1.jpg}
}

\begin{figure}[t]
    \centering 
    \begin{tabular}{@{}c@{}c@{}c@{}}
        \depthmapbrandenburgpic{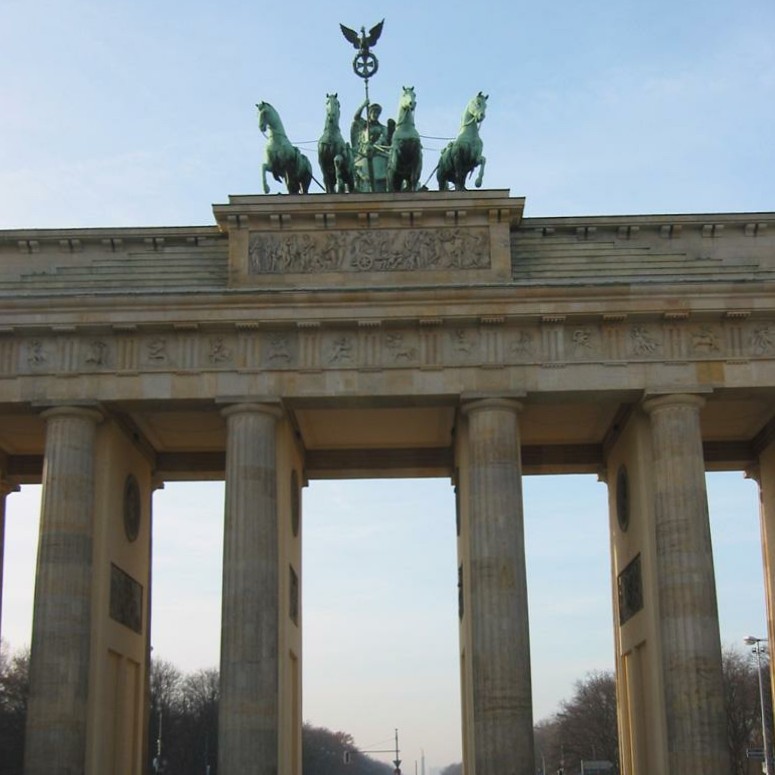} &
        \depthmapbrandenburgpic{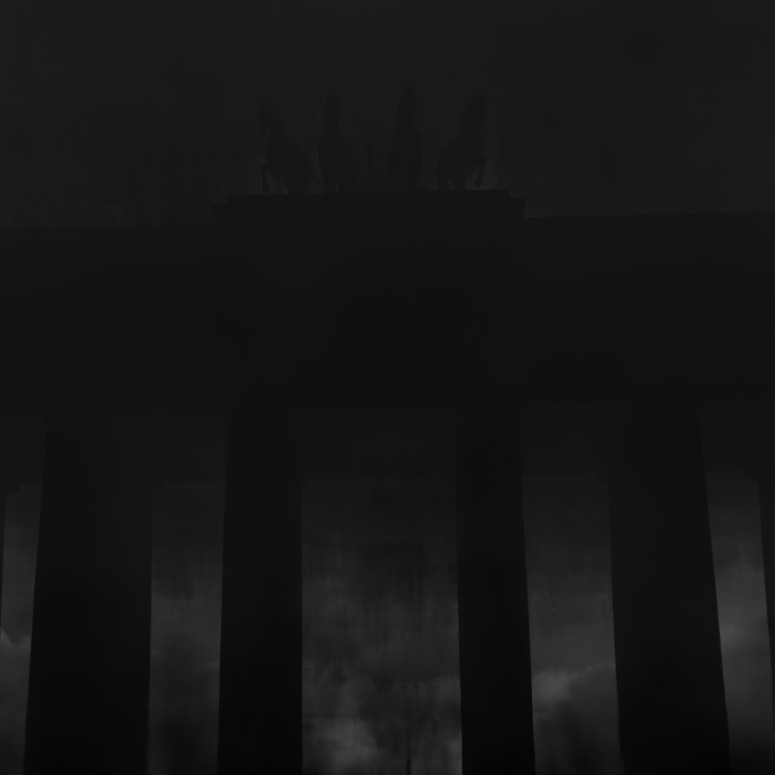} &
        \depthmapbrandenburgpic{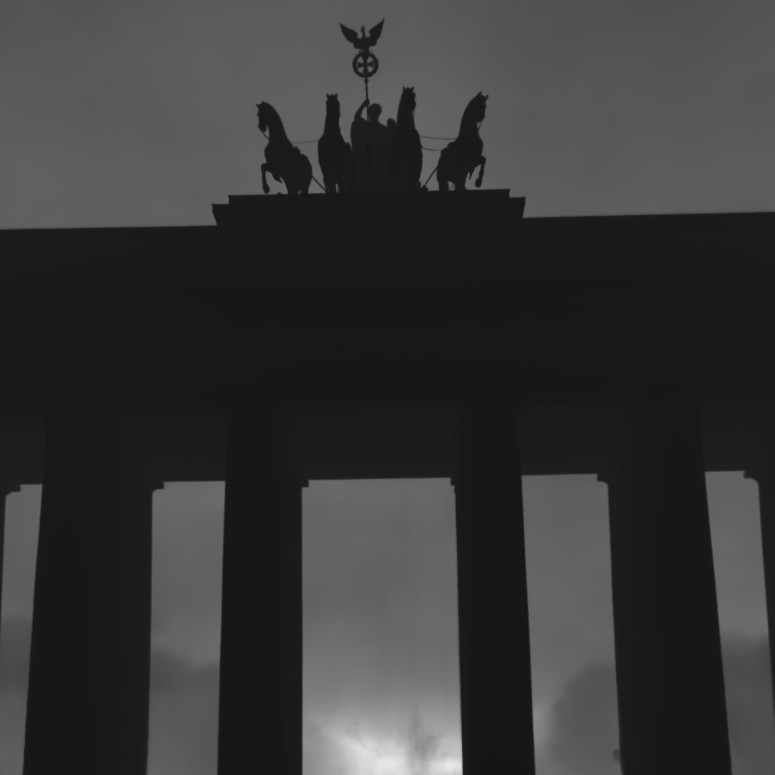}\\
        \depthmapsacrepic{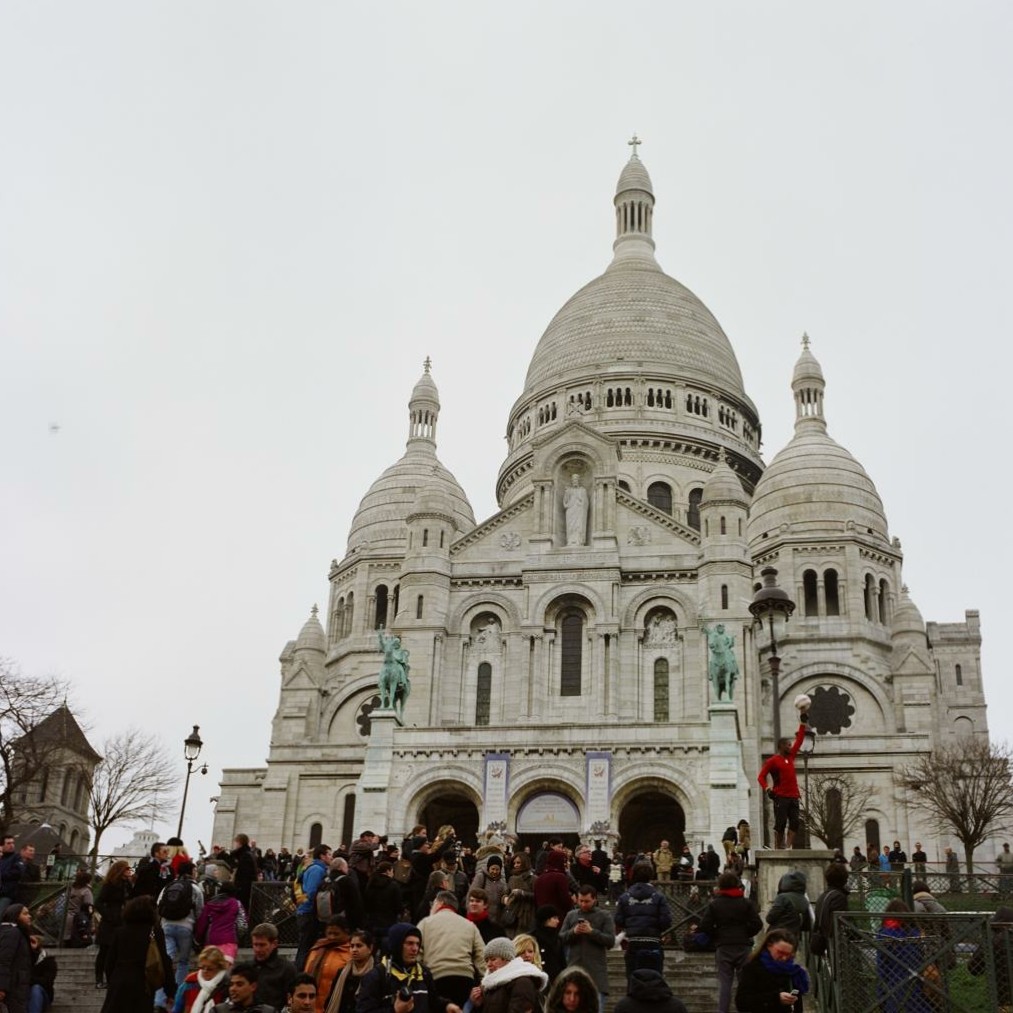} &
        \depthmapsacrepic{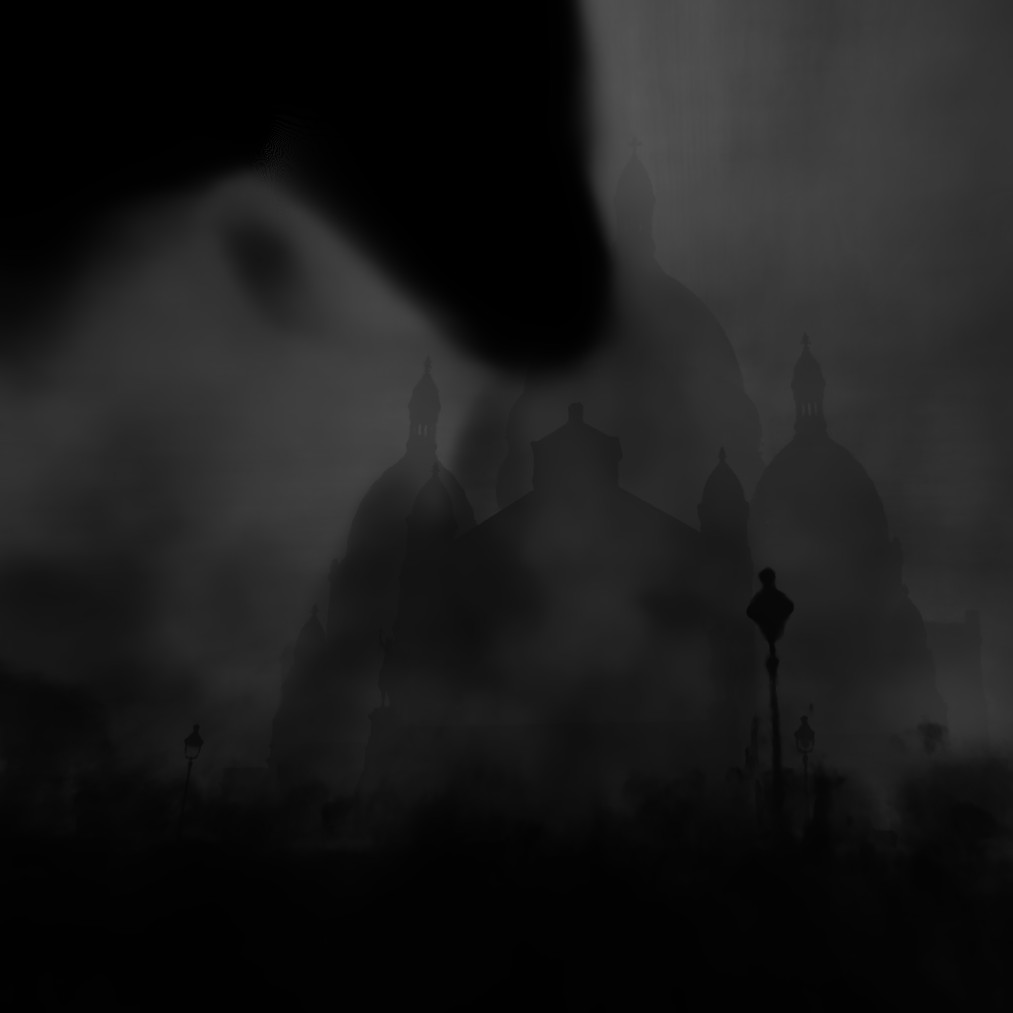} &
        \depthmapsacrepic{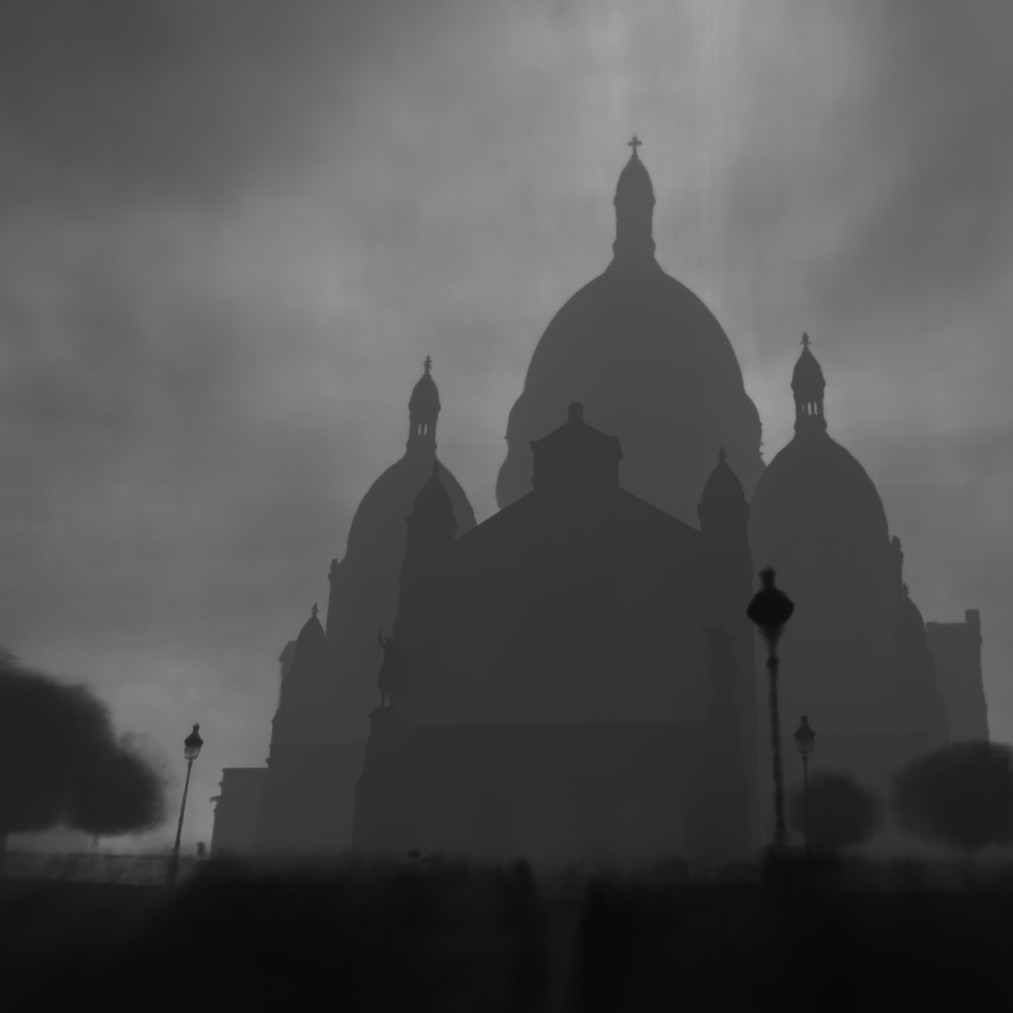}
        \\
        \small (a) Reference &
        \small (b) \NERF{} &
        \small (c) \NERFW{}
    \end{tabular} 
    \caption{
        Depth maps from \NERF{} and \NERFW{}, rendered by computing the expected termination depth of each ray.
        \NERF{}'s geometry is corrupted by appearance variation and occluders, while \NERFW{} is robust to such phenomena and produces accurate 3D reconstructions.
        \flickrcc{burkeandhare, photogreuhphies}.
    }
    \label{fig:depth-maps}
\end{figure}

\newcommand\brandenburgpic[1]{
    \raisebox{-0.5\height}{
    \begin{tikzpicture}[
    zoomboxarray,
    zoomboxes below,
    connect zoomboxes,
    zoombox paths/.append style={ultra thick}]
        \node[image node]{\includegraphics[width=0.154\textwidth]{brandenburg/v2/#1.jpg}};
        \zoombox[magnification=4,color code=nicered]{0.48,0.86}
        \zoombox[magnification=4,color code=niceblue]{0.54,0.55}
    \end{tikzpicture}}
}

\newcommand\sacrepic[1]{
    \raisebox{-0.5\height}{
    \begin{tikzpicture}[
    zoomboxarray,
    zoomboxes below,
    connect zoomboxes,
    zoombox paths/.append style={ultra thick}]
        \node[image node]{\includegraphics[width=0.154\textwidth]{sacre/v2/#1.jpg}};
        \zoombox[magnification=4,color code=nicered]{0.68,0.18}
        \zoombox[magnification=3,color code=niceblue]{0.32,0.9}
    \end{tikzpicture}}
}

\newcommand\trevipic[1]{
    \raisebox{-0.5\height}{
    \begin{tikzpicture}[
    zoomboxarray,
    zoomboxes below,
    connect zoomboxes,
    zoombox paths/.append style={ultra thick}]
        \node[image node]{\includegraphics[width=0.154\textwidth]{trevi/v2/#1.jpg}};
        \zoombox[magnification=5,color code=nicered]{0.681,0.595}
        \zoombox[magnification=6,color code=niceblue]{0.715,0.495}  
    \end{tikzpicture}}
}

\newcommand\praguepic[1]{
    \raisebox{-0.5\height}{
    \begin{tikzpicture}[
    zoomboxarray,
    zoomboxes below,
    connect zoomboxes,
    zoombox paths/.append style={ultra thick}]
        \node[image node]{\includegraphics[width=0.154\textwidth]{prague/#1.jpeg}};
        \zoombox[magnification=6,color code=nicered]{0.235,0.170}  
        \zoombox[magnification=6,color code=niceblue]{0.560,0.555}
    \end{tikzpicture}}
}

\newcommand\tajpic[1]{
    \raisebox{-0.5\height}{
    \begin{tikzpicture}[
    zoomboxarray,
    zoomboxes below,
    connect zoomboxes,
    zoombox paths/.append style={ultra thick}]
        \node[image node]{\includegraphics[width=0.154\textwidth]{taj/#1.jpeg}};
        \zoombox[magnification=5,color code=nicered]{0.431,0.83}
        \zoombox[magnification=6,color code=niceblue]{0.490,0.640}  
    \end{tikzpicture}}
}

\newcommand\hagiapic[1]{
    \raisebox{-0.5\height}{
    \begin{tikzpicture}[
    zoomboxarray,
    zoomboxes below,
    connect zoomboxes,
    zoombox paths/.append style={ultra thick}]
        \node[image node]{\includegraphics[width=0.154\textwidth]{hagia/#1.jpeg}};
        \zoombox[magnification=5,color code=nicered]{0.623,0.950}  
        \zoombox[magnification=6,color code=niceblue]{0.757,0.107} %
    \end{tikzpicture}}
}

\begin{figure*}

    \centering 
    \setlength{\tabcolsep}{-1.5pt}
    \def\arraystretch{1}
    \begin{tabular}{c@{\hskip 2mm}cccccc}
    \rotatebox[origin=c]{90}{\hspace{10mm} Prague Old Town} &
        \praguepic{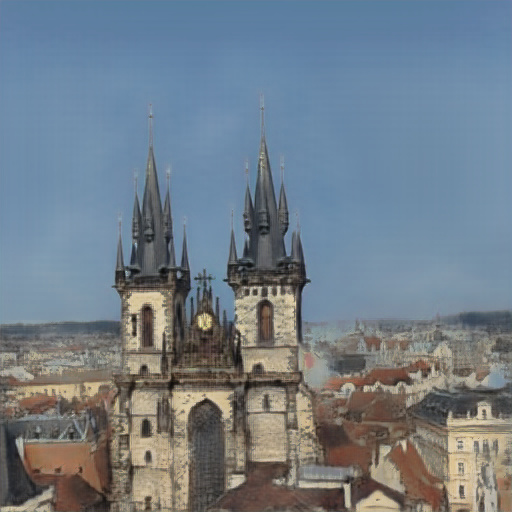} &
        \praguepic{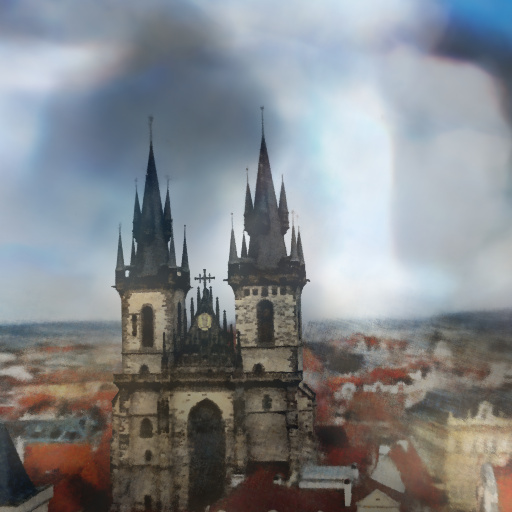} &
        \praguepic{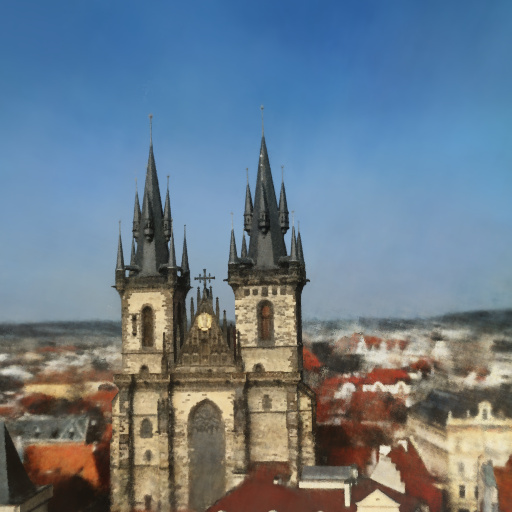} &
        \praguepic{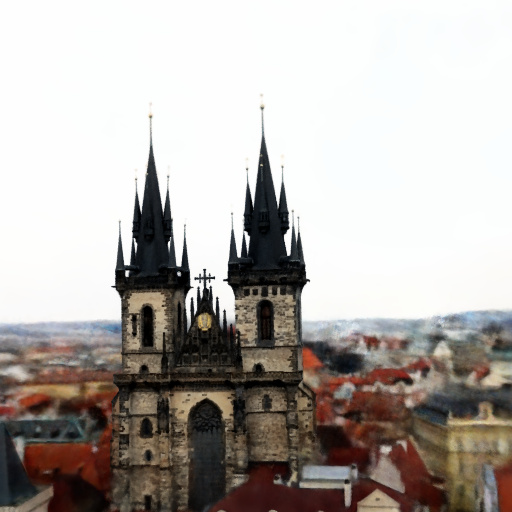} &
        \praguepic{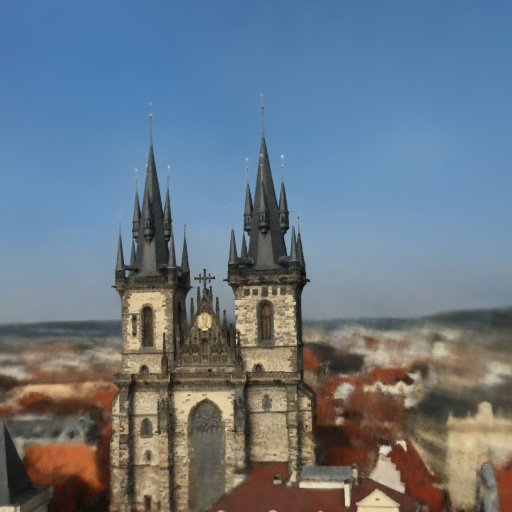} &
        \praguepic{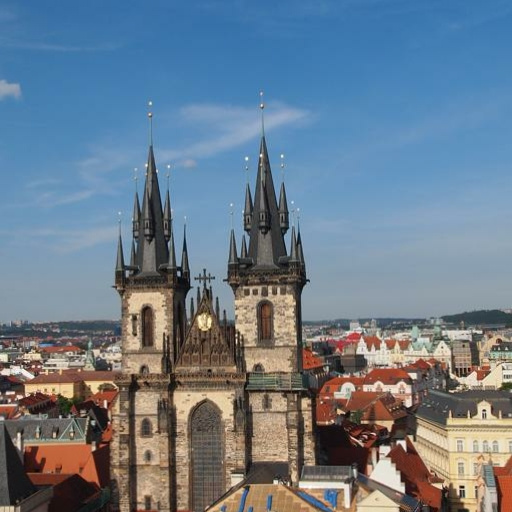}
    \\ [-11mm]
    \rotatebox[origin=c]{90}{\hspace{10mm} Sacre Coeur} &
        \sacrepic{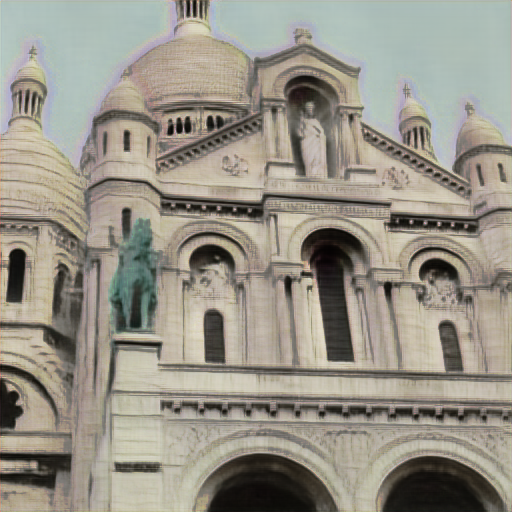} &
        \sacrepic{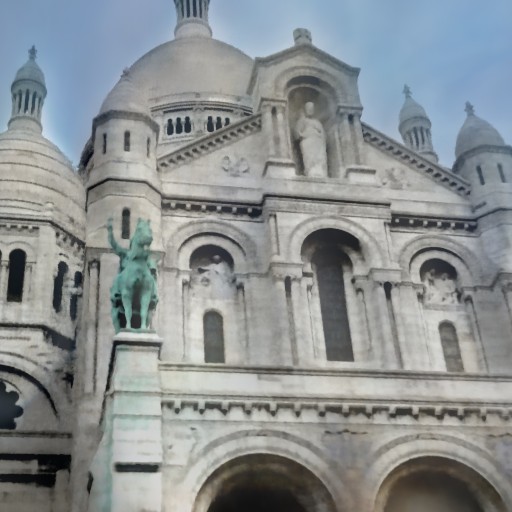} &
        \sacrepic{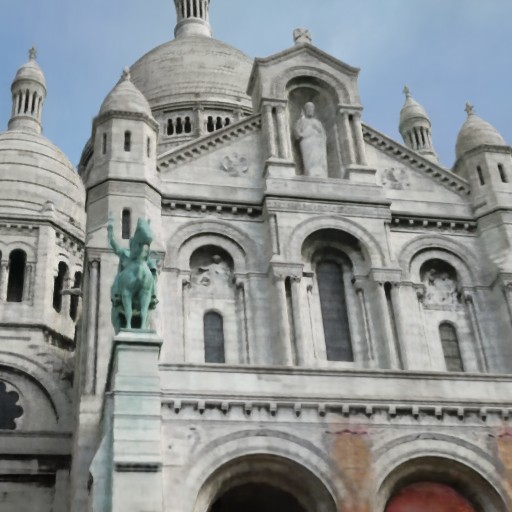} &
        \sacrepic{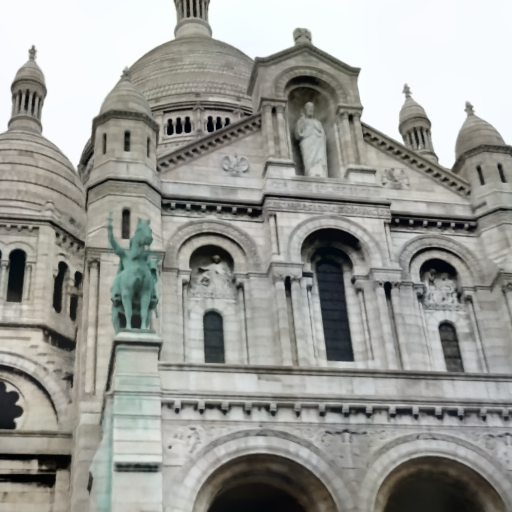} &
        \sacrepic{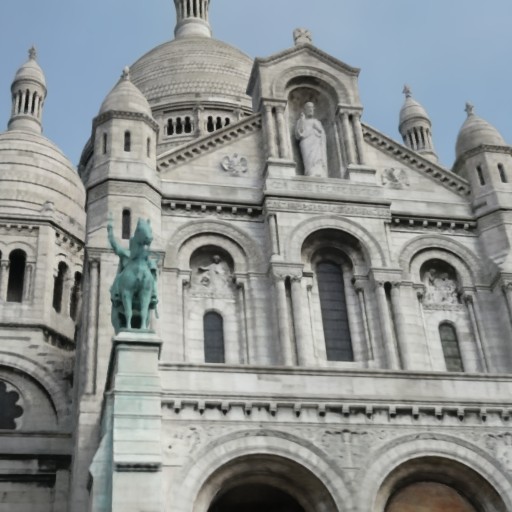} &
        \sacrepic{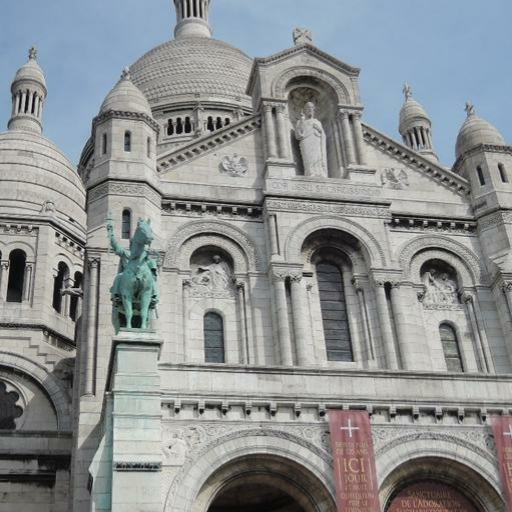}
    \\ [-11mm]
    \rotatebox[origin=c]{90}{\hspace{10mm} Taj Mahal} &
        \tajpic{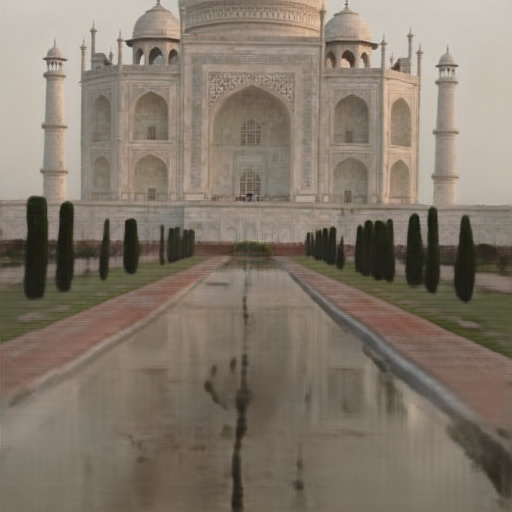} &
        \tajpic{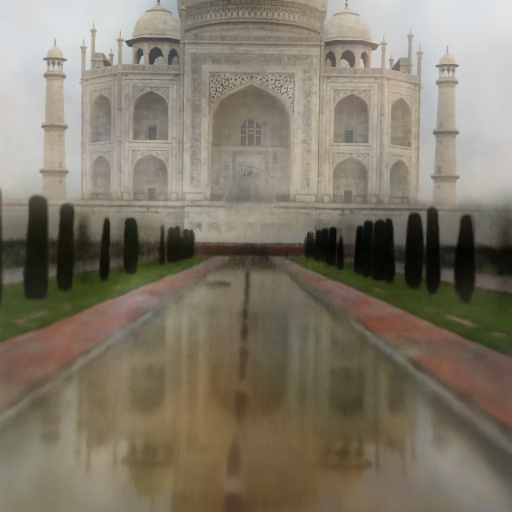} &
        \tajpic{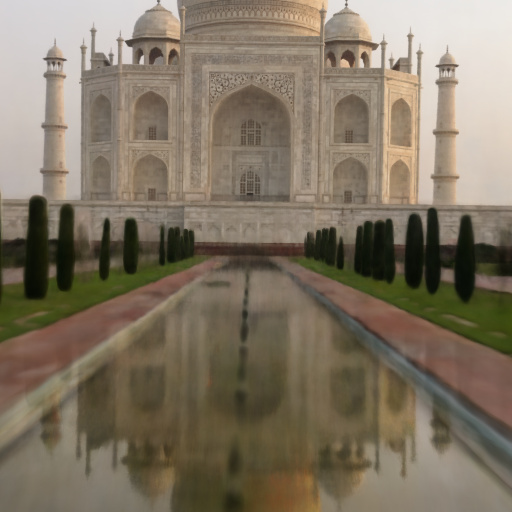} &
        \tajpic{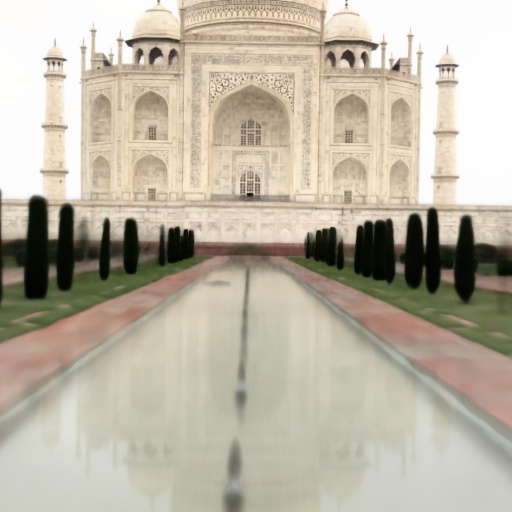} &
        \tajpic{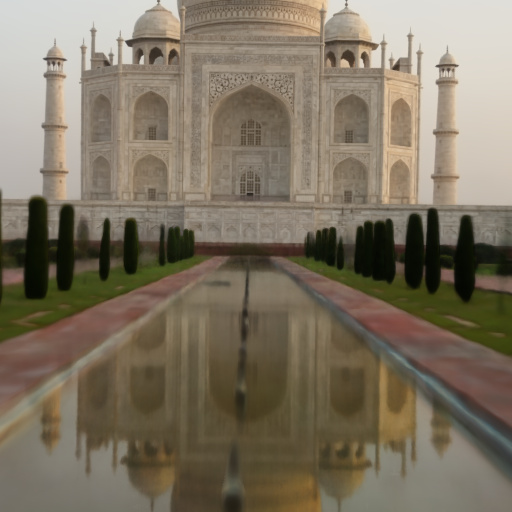} &
        \tajpic{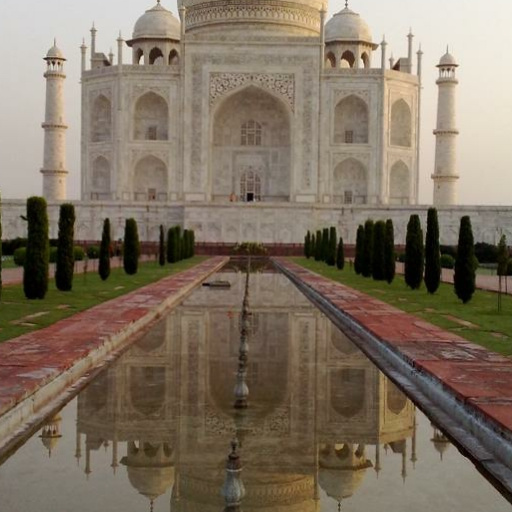}
    \\ [-11mm]
    &
    \multicolumn{1}{c}{\small NRW} &
    \multicolumn{1}{c}{\small \NERF{}} &
    \multicolumn{1}{c}{\small \NERFG{}} &
    \multicolumn{1}{c}{\small \NERFU{}} &
    \multicolumn{1}{c}{\small \NERFW{}} &
    \multicolumn{1}{c}{\small Ground-truth}
    \\
    \end{tabular} 
    \caption{
        Qualitative results from experiments on the Phototourism dataset.
        \NERFW{} is simultaneously able to model appearance variation (top), remove transient occluders (flag, middle), and reconstruct fine details in the scene (bottom).
        Further datasets are shown in Figure~\ref{fig:phototourism_qualitative_2} (supplementary).
        \flickrcc{firewave, clintonjeff, leoglenn\_g}.
    } 
    \label{fig:phototourism_qualitative}
\end{figure*}

\newcommand{\glopic}[1]{
    \adjincludegraphics[max width=0.14\textwidth,max height=0.095\textwidth]{glo_interp/brandenburg/#1.jpg}
}

\begin{figure*}[h!]
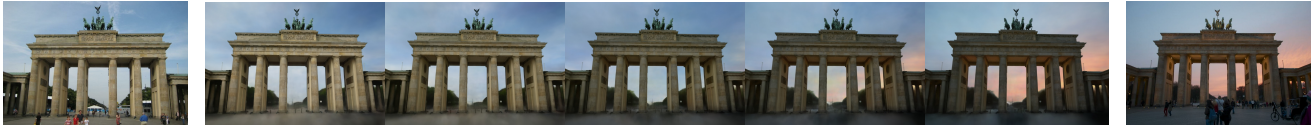

    \centering 
    \setlength{\tabcolsep}{-2pt}
    \def\arraystretch{1}
    \begin{tabular}{c@{\hskip 1.5mm}ccccc@{\hskip 1.5mm}c}
        \glopic{000113} &
        \glopic{000020_rgb} &
        \glopic{000022_rgb} &
        \glopic{000025_rgb} &
        \glopic{000027_rgb} &
        \glopic{000029_rgb} &
        \glopic{000227}
    \end{tabular}
    \vspace{-0.1in}
    \caption{
        Interpolations between the appearance embeddings $\glo{\embed}$ of two training images (left, right), which results in renderings (middle) where color and illumination are interpolated but geometry is fixed.
        Note that the training images contain people (left) and lights (right) that do not appear in the renderings.
        \flickrcc{mightyohm, blatez}.
    }
    \label{fig:glo-interp}
\end{figure*}

\minisection{Results:}
Figure~\ref{fig:phototourism_qualitative} shows qualitative results for all models and baselines on a subset of scenes. 
NRW produces renderings with checkerboard artifacts characteristic of 2D re-rendering methods~\cite{johnson2016perceptual}.
NRW is also sensitive to upstream errors in 3D geometry such as incomplete point clouds, as can be seen in the smaller towers of the church in the Prague Old Town.
\NERF{} produces a consistent 3D geometry, but large parts of the scene have ghosting artifacts and occlusions, which are particularly noticeable on Sacre Coeur and Prague Old Town.
Renderings from \NERF{} also tend to exhibit strong global color shifts when compared to the ground truth.
These artifacts are the direct consequence of \NERF{}'s static-world assumption --- \NERF{} attempts to explain away all photometric variation and transient occlusion using a single scene representation.
This static assumption impairs not only  \NERF{}'s renderings but also its underlying geometry, while \NERFW{} produces accurate 3D reconstructions (\figref{fig:depth-maps}).

The \NERFG{} ablation produces less ``foggy'' renderings than \NERF{}, as shown in Figure~\ref{fig:phototourism_qualitative}.
However, \NERFG{} is unable to reconstruct high-frequency details such as the brickwork on Sacre Coeur's dome.
In contrast, the \NERFU{} ablation is better able to capture fine detail, but is unable to model varying photometric effects.
\NERFW{} has the benefits of both ablations, and thereby produces sharper and more accurate renderings.

Quantitative results are summarized in~\tabref{tab:phototourism_half}.
Optimizing \NERF{} on \inthewild{} photo collections leads to particularly poor results that are unable to compete with NRW.
In contrast, \NERFW{} outperforms the baselines on PSNR and MS-SSIM across all datasets.
In particular, \NERFW{} improves over the previous state of the art NRW by an average margin of 4.4dB in PSNR, and with up to 40\% improvements in MS-SSIM.
In spite of minimizing only a per-pixel squared error during training, \NERFW{} improves upon the prior state of the art on LPIPS in 3 of 6 scenes and remains competitive in the remainder.
Lacking a perceptual loss, \NERFW{} is not incentivized to produce the high-frequency textures favored by perceptual metrics such as LPIPS.
However, NRW exhibits temporal instability --- as the camera moves, renderings appear to flicker and wobble unrealistically, and this is not captured by the single-image metrics or figures used in this paper. 
We strongly encourage the reader to inspect the supplemental video to observe the temporal instability of NRW compared to \NERF{} and \NERFW{}.

\minisection{Controllable Appearance:}

\newcommand{\videoframespic}[1]{
    \includegraphics[trim=0 0 0 240, clip, height=22mm]{epipolar/#1}
}

\newcommand\epipic[1]{
    \includegraphics[trim=0 0 0 0, clip, height=22mm]{epipolar/brandenburg_gate_v2/#1}
}

\begin{figure*}
    \centering 
    \small
    \setlength{\tabcolsep}{0pt}
    \def\arraystretch{1}
    \begin{tabular}{c@{\hskip 2mm}ccc}
        \videoframespic{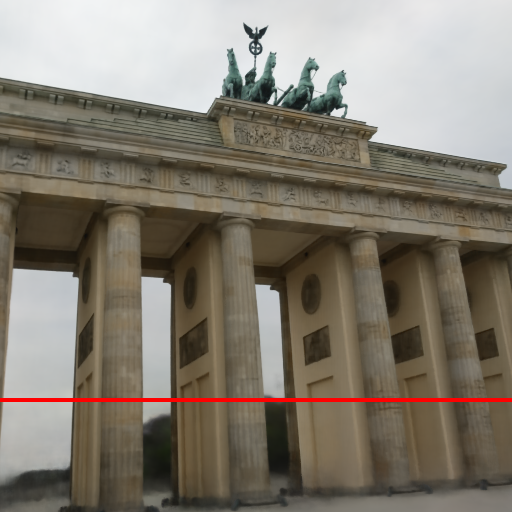} &
        \epipic{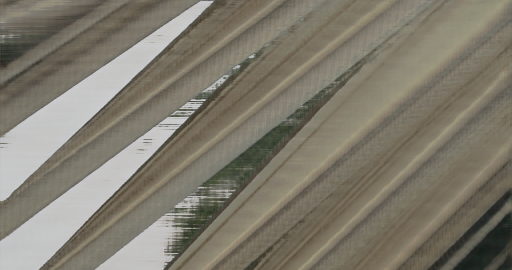} &
        \epipic{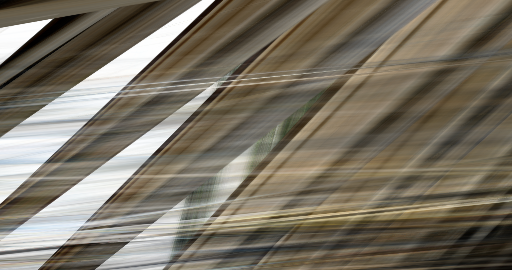} &
        \epipic{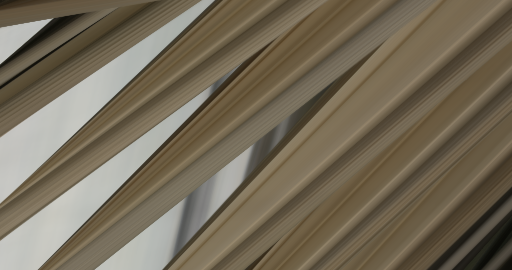} \\
        Reference &
        \textsc{NRW} &
        \textsc{\NERF{}} &
        \textsc{\NERFW{}} \\
    \end{tabular}
    \vspace{-2mm}
    \caption{
        Epipolar plane images (EPI) synthesized from videos rendered by different models for the Brandenburg Gate scene.
        The camera is translated from left to right along a straight path, and the horizontal line at the same position (red line, reference) is taken across all video frames and stacked vertically, producing the EPIs shown above.
        A temporally consistent video results in a clean and smooth EPI, while noise in an EPI indicates temporal flickering artifacts.
        NRW's video contains heavy flickering with transient objects popping in and out of the frame while \NERF{} produces severe ghosting artifacts in front of the landmarks.
        \NERFW{} produces highly temporally consistent videos.
        We strongly encourage the readers to watch the video in the supplementary material.
    }
    \label{fig:flythrough_frames}
\end{figure*}

One consequence of modeling appearance with a latent embedding space $\glo{\embed} \in \mathbb{R}^{\glo{n}}$ is that it enables the modification of lighting and appearance of a rendering without altering the underlying 3D geometry.
In Figure~\ref{fig:cover_figure} (right), we see slices of four rendered images produced by \NERFW{} using appearance embeddings associated with four training set images.
In addition to the embeddings associated with images in the training set, one may also apply \NERFW{} to arbitrary vectors in the same space.
In~\figref{fig:glo-interp}, we present five images rendered from a fixed camera position, where we interpolate between the appearance embeddings associated with the left and right training images.
Note that the appearance of the rendered images smoothly transitions between the two end points without introducing artifacts to the 3D geometry.
We encourage readers to view the supplementary video to better appreciate the naturalness of such interpolations.

\minisection{View-consistency:}
\figref{fig:flythrough_frames} shows ``flatland'' light field renderings for NRW, \NERF{}, and \NERFW{} with the camera panning along a straight path.
Renderings from \NERFW{} are more view-consistent (the Lambertian scene content is correctly reconstructed as being constant across viewing directions) and exhibits significantly less flickering than NRW or \NERF{}.
NRW is unable to model temporal consistency between frames for transient objects, while \NERF{} is forced to embed view-dependent effects as colored fog into its scene representation.

\newcommand{\limitationspictwo}[1]{
    \raisebox{-.5\height}{
        
    }
}

\begin{figure}[b!]
    \centering 
    \setlength{\tabcolsep}{2pt}
    \begin{tabular}{cc}
        \includegraphics[width=0.48\columnwidth]{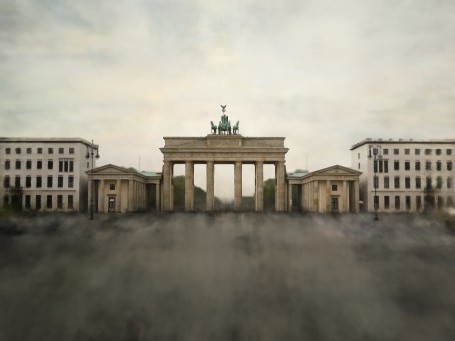} &
        \includegraphics[width=0.48\columnwidth]{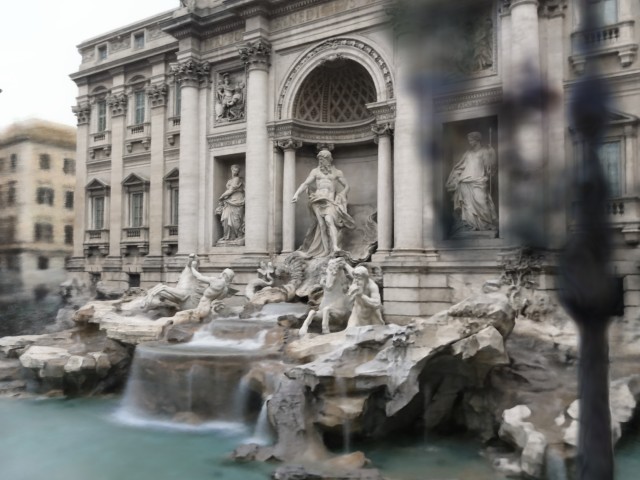} \\
    \end{tabular} 
    \vspace{-0.1in}
    \caption{Limitations of \NERFW{} on the Phototourism dataset. Rarely-seen parts of the scene (ground, left) and incorrect camera poses (lamp post, right) can result in blur.
    }
    \label{fig:limitations2}
\end{figure}

\minisection{Limitations:}
While \NERFW{} is able to produce photorealistic and temporally consistent renderings from unstructured photographs, rendering quality degrades in areas of the scene that are rarely observed in the training images, or only observed at very oblique angles, like the ground, as shown in Figure~\ref{fig:limitations2}.
Similar to \NERF{}, \NERFW{} is also sensitive to camera calibration errors, which can lead to blurry reconstructions on the parts of the scene that have been imaged by incorrectly-calibrated cameras.

\minisection{Synthetic Experiments:}
The components of \NERFW{} were designed to deal with specific forms of photometric inconsistency, such as color shifts and occluders. Unfortunately, the uncontrolled nature of the Phototourism dataset means that it is challenging to demonstrate that each model component does indeed address the confounding factor that it was designed to address. For this reason, in the supplement we present a controlled ablation study in which we construct variations of a synthetic dataset used in~\cite{mildenhall2020nerf} wherein we manually introduce the phenomena we expect to find in \inthewild{} imagery. As can be seen in the supplement, the results of this ablation study are consistent with our expectations.

\section{Conclusion} \label{sec:conclusion}

We have presented \NERFW{}, a novel approach for 3D scene reconstruction of complex environments from unstructured internet photo collections that builds upon \NERF{}.
We learn a per-image latent embedding capturing photometric appearance variations often present in \inthewild{} data, and we decompose the scene into image-dependent and shared components to allow our model to disentangle transient elements from the static scene. 
Experimental evaluation on real-world (and synthetic) data demonstrates significant qualitative and quantitative improvement over previous state-of-the-art approaches.

{\small
\bibliographystyle{ieee_fullname}
\bibliography{bib}
}

\clearpage
\appendix

\section*{Supplemental Materials}

We {\bf strongly} encourage the reader to view the video results included in the supplemental materials.

\newcommand{\filterspic}[1]{
    \adjincludegraphics[width=0.49\columnwidth]{filtered/#1.jpg}
}

\begin{figure}[b]
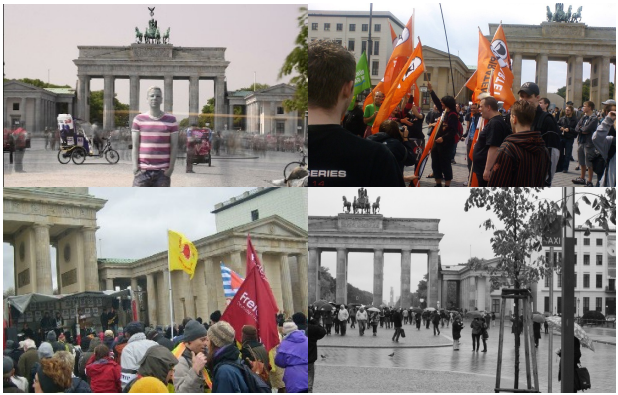

    \centering 
    \setlength{\tabcolsep}{-0.5pt}
    \def\arraystretch{1}
    \begin{tabular}{c@{\!\!}c}
        \filterspic{ex_2} &
        \filterspic{ex_10}
        \\ [-4mm]
        \filterspic{ex_5} &
        \filterspic{ex_8}
    \\
    \end{tabular} 
    \caption{
        Samples of images removed by pre-processing. Images where transient objects occupy more than 80\% of the image or where the NIMA score is below a threshold are discarded from the scene.
        \flickrcc{alcanthus, headnut, uwehiksch, and stevebaty}.
    }
    \label{fig:filtered-examples}
\end{figure}

\section{Model Parameters}
\label{app:hparams}
In the following, we document the selected hyperparameters for \NERFW{}. 
A grid search over the hyperparameter space was employed to select the below values. 
Note that the values were optimized for the Brandenburg Gate scene of the Phototourism dataset and reused for the remaining scenes. 
As in \NERF{}, we employ an MLP architecture for modeling density and radiance. 
We apply $8$ layers of $512$ hidden units for the `Base' component, and $4$ layers of $128$ hidden units for both `Static` and ``Transient / Uncertainty' components; see Figure~\ref{fig:model-arch}).
Regarding our positional encoding functions $\gamma$ we use $15$ frequencies for when encoding position and $4$ when encoding viewing direction. 
During training, $512$ points per ray are sampled from each of the coarse and fine models for a total of $1024$ points per ray. 
We double that number during evaluation to \num[group-separator={,}]{2048}. 
The latent embedding vector for appearance has an embedding dimension of size $\glo{n}=48$. 
For transient objects, we use an embedding dimension of size $\uncertainty{n}=16$. 
We select a value of $0.01$ for the $L_1$ regularizer multiplier $\lambda_{u}$ and $0.03$ as the minimum importance $\beta_{\min}$.
Models are trained with Adam over 300,000 iterations with a batch size of 2048 and an initial learning rate of $0.001$, decaying ten-fold every 150,000 iterations.

\section{Evaluation}

The problem of evaluating the performance of a view synthesis system on \inthewild{} photo collections is itself a challenging problem, as it reduces to the still-open problem of measuring perceptual image similarity~\cite{pappas2000perceptual, thung2009survey, wang2002universal, zhang2018unreasonable}.
Though LPIPS~\cite{zhang2018unreasonable} is an acceptable solution in many contexts, 
two perfectly-aligned images of the same 3D structure imagined at different times of day or under different lighting conditions generally result in significant ``perceptual'' image differences.
This presents a challenge when evaluating \NERFW{} (and its ablation \NERFG{}). 
After training, only images in the training set are assigned optimized appearance embeddings $\glo{\embed}$ --- our model does not recover a single model of the world, it recovers a \emph{family} of solutions under a variety of appearances. 
To evaluate a test-set image, we must therefore identify the image's corresponding $\glo{\embed}$ embedding. 

Naive solutions to this problem, such as setting $\glo{\embed}$ to a vector of zeros or to the mean of optimized embeddings, results in plausible renderings that, while structurally accurate, fail to match the appearance of the ground truth.
As a result, perceptual metrics are unable to credibly measure the quality of each method's scene representation.
To address this deficiency, we evaluate \NERFG{} and \NERFW{} by optimizing appearance embedding $\glo{\embed}$ on the \emph{left half} of each ground-truth image and calculating metrics on the corresponding \emph{right half}. 
This split-image evaluation scheme enables \NERFG{} and \NERFW{} to adapt the scene's appearance without directly optimizing held-out pixels.
Note that by virtue of the model's design, changes to the appearance embedding \emph{cannot} alter the underlying scene's geometry (see Figures~\ref{fig:phototourism_qualitative} and \ref{fig:glo-interp}), further limiting the potential for information leakage.

Finally, note that the NRW baseline captures appearance by encoding the \emph{entirety} of a held-out image with an appearance encoder model. 
Unlike \NERFW{}, this method is unable to isolate geometry from appearance and, given a sufficiently high-dimensional space, is capable of storing the image itself.
In preliminary experiments, we notice a small drop in performance when applying a similar split-image evaluation scheme as described above.
To remain comparable to prior results, we replicate the evaluation scheme as originally published.

For the LPIPS error metric we use the AlexNet implementation provided at \url{https://github.com/richzhang/PerceptualSimilarity}.

\section{Phototourism Dataset}
\label{app:phototoursim}

As a coarse pre-filtering step, we remove low quality images consisting largely of transient objects by omitting those with a NIMA~\cite{talebi2018nima} score below $3$. 
We further filter out images where transient objects occupy more than $80\%$ of the image's area according to a DeepLab v3 \cite{chen2017rethinking} model trained on Ade20k.
\figref{fig:filtered-examples} depicts some examples of the filtered images from the Brandenburg Gate scene.

For quantitative evaluation, we form a test set by hand-selecting photos representative of the qualities we intend to replicate: well-focused and without occluders.
While a naive random selection of images may seem appropriate, image comparison metrics such as PSNR, MS-SSIM, and LPIPS are unable to ignore transient objects.
Indeed, \NERFW{} is designed to generate images \emph{without} such occluders, and so will score poorly when evaluated on a reference image that contains occluders.
We therefore explicitly select photos without transient phenomena or extreme photometric effects.
Photos constituting the test set were chosen during the preliminary experiments stage and held-out until the final evaluation shown in \tabref{tab:phototourism_half}.
In particular, the chosen photos were not used to guide model design or hyperparameter search.
See \tabref{tab:phototourism-statistics} for scene-specific statistics on this dataset.

\begin{table}
    \footnotesize
    \begin{sc}
    \begin{tabular}{lrrrr}
        \toprule
        & \multicolumn{2}{c}{Train} & \multicolumn{2}{c}{Validation} \\
        \cmidrule(r){2-3}\cmidrule(r){4-5}
        Dataset & Images & Pixels & Images & Pixels \\
        \cmidrule(r){2-3}\cmidrule(r){4-5}
        Brandenburg Gate & 763 & 564M & 38 & 12M \\
        Sacre Coeur & 830 & 605M & 40 & 14M \\
        Trevi Fountain & 1,689 & 1249M & 39 & 14M \\
        Taj Mahal & 811 & 581M & 27 & 9M \\
        Prague & 2,000 & 1417M & 28 & 9M \\
        Hagia Sophia & 606 & 434M & 29 & 10M \\
        \bottomrule
    \end{tabular}
    \end{sc}
    \caption{
        Number of images and pixels per Phototourism scene. 
        Pixel counts measuring in millions.
    }
    \label{tab:phototourism-statistics}
\end{table}
\section{Lego Dataset}
\label{app:lego}

\newcommand{\colorperturbs}[1]{
    \adjincludegraphics[max width=0.285\columnwidth]{lego/perturbations/colors_#1.jpg}
}
\newcommand{\occluderperturbs}[1]{
    \adjincludegraphics[max width=0.285\columnwidth]{lego/perturbations/occluders_#1.jpg}
}
\begin{figure}[!b]
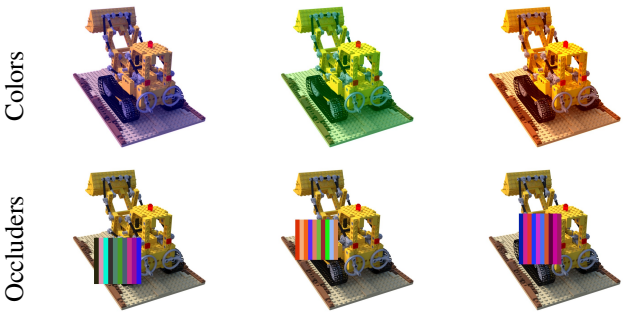

    \centering 
    \setlength{\tabcolsep}{-0.6pt}
    \def\arraystretch{1}
    \begin{tabular}{c@{\hskip 2mm}c@{\hskip 3mm}c@{\hskip 3mm}c@{\hskip 3mm}}
    \rotatebox[origin=c]{90}{\hspace{18mm} Colors} &
        \colorperturbs{1} &
        \colorperturbs{2} &
        \colorperturbs{3}
    \\ [-15mm]
    \rotatebox[origin=c]{90}{\hspace{18mm} Occluders} &
        \occluderperturbs{1} &
        \occluderperturbs{2} &
        \occluderperturbs{3}
    \\ [-15mm]
    \end{tabular} 
    \caption{Examples of perturbations applied to the Lego dataset. The top row shows various color perturbations applied to the same view, whereas the bottom shows the effect of randomly adding occluders to the same view.}
    \label{fig:perturbations-examples}
\end{figure}

For a controlled ablation study, we construct variants of the Lego dataset~\cite{mildenhall2020nerf} inspired by effects we expect to find \inthewild{} (Table~\ref{tab:lego}).

\newcommand{\colorscale}{s}
\newcommand{\coloroffset}{b}

\minisection{Color perturbations:}
To simulate variable lighting and exposure, we apply a random scale and shift transformation to the RGB values of each image.
In particular, we replace each training image $\image_i \in [0,1]^{800 \times 800 \times 3}$ with $\tilde{\image}_i$ where, for each RGB color channel $j$, $\tilde{\image}_{ij} = \min(1, \max(0, \colorscale_{ij} \image_{ij} + \coloroffset_{ij}))$ where scale $\colorscale_{ij} \sim \mathcal{U}(0.8, 1.2)$ and offset $\coloroffset_{ij} \sim \mathcal{U}(-0.2, 0.2)$ are sampled uniformly at random for each $i$ and $j$.
Qualitatively, this results in variable tint and brightness.
We apply perturbations to all training images except the first, whose appearance embedding is used to render novel views. The top row of~\figref{fig:perturbations-examples} shows the effect of applying random color perturbations to the same image.

\minisection{Occlusions:}
We simulate transient occluders by drawing randomly-positioned and randomly-colored squares on each training image.
Each square consists of ten vertical, colored stripes with colors chosen at random.
Like transient occluders in the real world, these squares do not have a consistent 3D location from image to image.
We again leave the first training image untouched for reference. \figref{fig:perturbations-examples} shows the effect of adding occlusions randomly to the same view.

\subsection{Experiments}
\label{app:lego-exp}

For the Lego datasets, we optimize for \num[group-separator={,}]{125000}
steps on 4 GPUs, which takes approximately 8 hours.
We use the same \NERF{} hyperparameters reported by Mildenhall et al.~\cite{mildenhall2020nerf} for all \NERF{} variants.
We optimize models on 100 images and evaluate on an additional 200, using the same \NERF{} model hyperparameters presented in the original work.
Hyperparameters specific to \NERFW{} (those not present in \NERF{}) are tuned for each dataset variation via grid search.

\minisection{Original:}
We begin by applying all methods to the original, unperturbed Lego dataset.
Quantitatively, we find that all model variations perform similarly (Table~\ref{tab:lego}).
While \NERF{} achieves slightly higher PSNR than all \NERFW{} variants, all other metrics suggest indistinguishable model quality.
We find that our implementation of \NERF{} performs slightly better than the performance reported in the original \NERF{} paper~\cite{mildenhall2020nerf}.

\minisection{Color Perturbations:}
\label{sec:lighting-variation}

We find that this change alone decreases \NERF{}'s PSNR by approximately 10dB on average (Table~\ref{tab:lego}).
As illustrated in Figure~\ref{fig:lego_qualitative}, \NERF{} is unable to isolate image-dependent photometric effects from its shared scene representation and thus entangles color variation with viewing direction.
\NERFG{} and \NERFW{}, on the other hand, isolate tinting using the appearance embedding $\glo{\embed}$.
Novel views rendered with a fixed appearance embedding demonstrate consistent color from all camera angles.
Quantitatively, we find both methods maintain almost identical metrics to those achieved on the original dataset.

\begin{table*}[t!]
\centering
    \small
    \setlength{\tabcolsep}{4pt}
    \def\arraystretch{1}
    \begin{sc}\begin{tabular}{l|ccc|ccc}
	&\multicolumn{3}{c|}{Original}&\multicolumn{3}{c}{Color Perturbations}\\
	&{\footnotesize$\uparrow$~PSNR}&{\footnotesize$\uparrow$~MS-SSIM}&{\footnotesize$\downarrow$~LPIPS}&{\footnotesize$\uparrow$~PSNR}&{\footnotesize$\uparrow$~MS-SSIM}&{\footnotesize$\downarrow$~LPIPS}\\
	\hline
	\NERF{}&\BEST{33.35}$\pm$0.05&\BEST{0.989}$\pm$0.000&\BEST{0.019}$\pm$0.000&23.38$\pm$0.05&0.964$\pm$0.001&0.076$\pm$0.001\\
	\NERFG{}&33.04$\pm$0.06&\BEST{0.989}$\pm$0.000&\SECOND{0.020}$\pm$0.000&\SECOND{30.66}$\pm$1.38&\SECOND{0.983}$\pm$0.007&\SECOND{0.031}$\pm$0.015\\
	\NERFU{}&\SECOND{33.07}$\pm$0.27&\BEST{0.989}$\pm$0.001&\BEST{0.019}$\pm$0.001&24.87$\pm$0.52&0.968$\pm$0.000&0.063$\pm$0.007\\
	\NERFW{}&32.89$\pm$0.14&\BEST{0.989}$\pm$0.000&\SECOND{0.020}$\pm$0.001&\BEST{31.51}$\pm$0.28&\BEST{0.987}$\pm$0.001&\BEST{0.022}$\pm$0.001\\
	\multicolumn{7}{c}{\,} \\
	&\multicolumn{3}{c|}{Occluders}&\multicolumn{3}{c}{Colors Perturbations \& Occluders}\\
	&{\footnotesize$\uparrow$~PSNR}&{\footnotesize$\uparrow$~MS-SSIM}&{\footnotesize$\downarrow$~LPIPS}&{\footnotesize$\uparrow$~PSNR}&{\footnotesize$\uparrow$~MS-SSIM}&{\footnotesize$\downarrow$~LPIPS}\\
	\hline
	\NERF{}&19.35$\pm$0.11&0.891$\pm$0.001&0.112$\pm$0.001&15.73$\pm$3.13&0.804$\pm$0.109&0.217$\pm$0.100\\
	\NERFG{}&22.71$\pm$0.63&0.922$\pm$0.005&0.086$\pm$0.003&\SECOND{21.08}$\pm$0.41&\SECOND{0.903}$\pm$0.007&\SECOND{0.116}$\pm$0.016\\
	\NERFU{}&\SECOND{23.47}$\pm$0.50&\SECOND{0.944}$\pm$0.004&\BEST{0.059}$\pm$0.004&17.65$\pm$4.10&0.846$\pm$0.130&0.183$\pm$0.117\\
	\NERFW{}&\BEST{25.03}$\pm$1.00&\BEST{0.946}$\pm$0.009&\SECOND{0.063}$\pm$0.009&\BEST{22.19}$\pm$0.30&\BEST{0.927}$\pm$0.003&\BEST{0.087}$\pm$0.004\\
	\end{tabular}\end{sc}
    \caption{
    Quantitative evaluation of \NERF{} and our proposed extensions on the synthetic Lego dataset.
        We report mean $\pm$ standard deviation across 5 independent runs with different random initializations.
        \BEST{Best} and \SECOND{second best} results are highlighted.
        On the {\sc Original} dataset, all models perform near identically.
        \NERF{} fails to varying degrees on the perturbed datasets because it has no mechanism to account for those perturbations.
        As expected, \NERFU{} fails on {\sc Colors}, but improves over \NERF{} on {\sc Occluders}.
        Likewise, \NERFG{} performs well on {\sc Colors} but fails on {\sc Occluders}.
        \NERFW{} is the only model that handles both types of perturbations.
    }
    \label{tab:lego}
\end{table*}

\minisection{Occluders:}

As shown in~\tabref{tab:lego}, this variation reduces \NERF{}'s PSNR by 14dB on average.
To reduce training error, \NERF{} and \NERFG{} represent occluders as colored fog in 3D space, thereby causing the Lego figure to be obscured (Figure~\ref{fig:lego_qualitative}).
While latent appearance embeddings were not designed to capture transient objects, we find that they enable \NERFG{} to reduce error by learning a radiance field that imitates the color of the underlying 3D geometry.
\NERFG{} and \NERFW{} are better able to isolate transient occluders from the static scene than their counterparts.

\minisection{Color Perturbations and Occluders:}
When both color and occluder perturbations are simultaneously enabled, we observe a decrease in performance across all methods, with \NERFW{} outperforming all baselines.
We further observe significant variation in model accuracy for both \NERF{} and \NERFU{} across five random seeds.
Both methods are poorly equipped to cope with photometric effects and occasionally fail to model the scene at all.

\newcommand\legopic[1]{
    \raisebox{-0.5\height}{
    \begin{tikzpicture}[
    zoomboxarray, 
    zoomboxes below, 
    connect zoomboxes,
    zoombox paths/.append style={ultra thick},
    zoomboxarray heightmultiplier=0.3]
        \node[image node]{\includegraphics[trim=152 82 52 85,clip,width=0.15\linewidth]{lego/#1.jpg}};
        \zoombox[magnification=4,color code=nicered]{0.30,0.55}
        \zoombox[magnification=3,color code=niceblue]{0.08,0.45}   %
    \end{tikzpicture}}
}

\begin{figure*}[!p]
    \centering 
    \setlength{\tabcolsep}{0pt}
    \begin{tabular}{@{}c@{\hskip 2mm}c@{\hskip 5mm}@{\hskip 5mm}c@{\hskip 3mm}c@{\hskip 3mm}c@{\hskip 3mm}c@{}}
    \rotatebox[origin=c]{90}{\hspace{18mm} Original} &
        \legopic{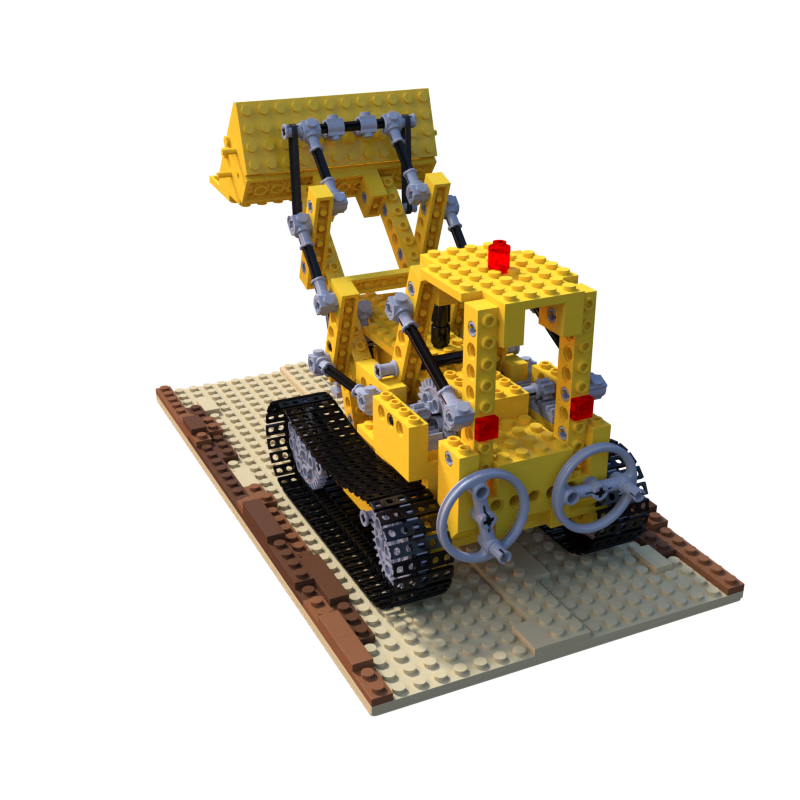} &
        \legopic{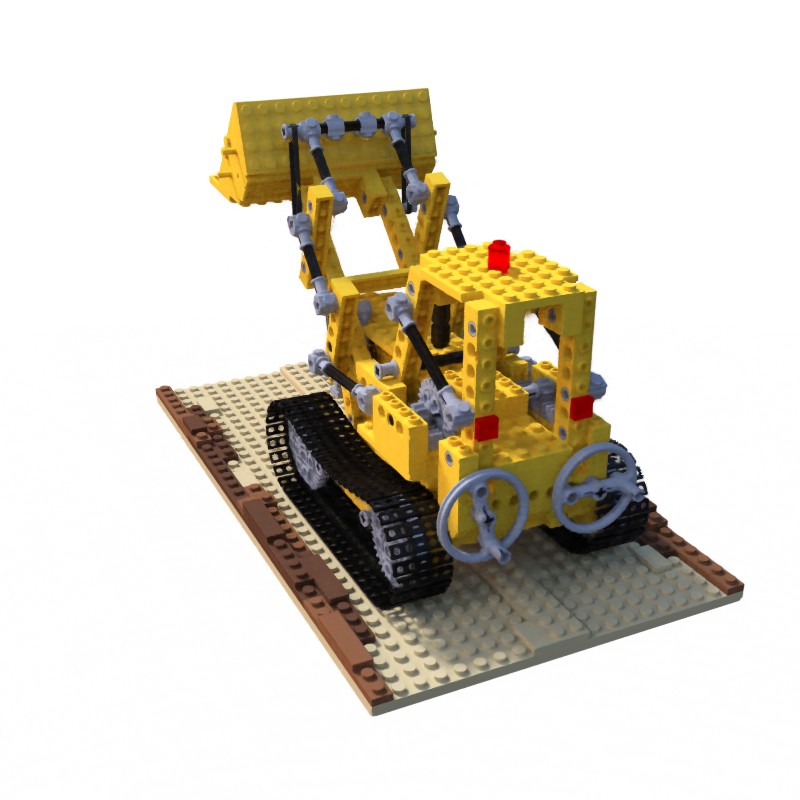} &
        \legopic{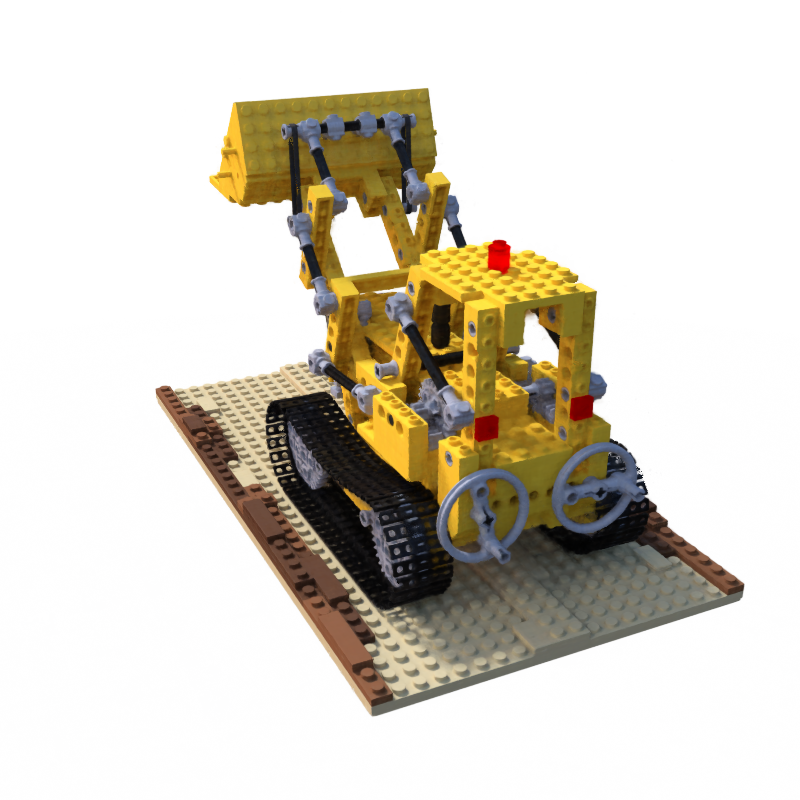} &
        \legopic{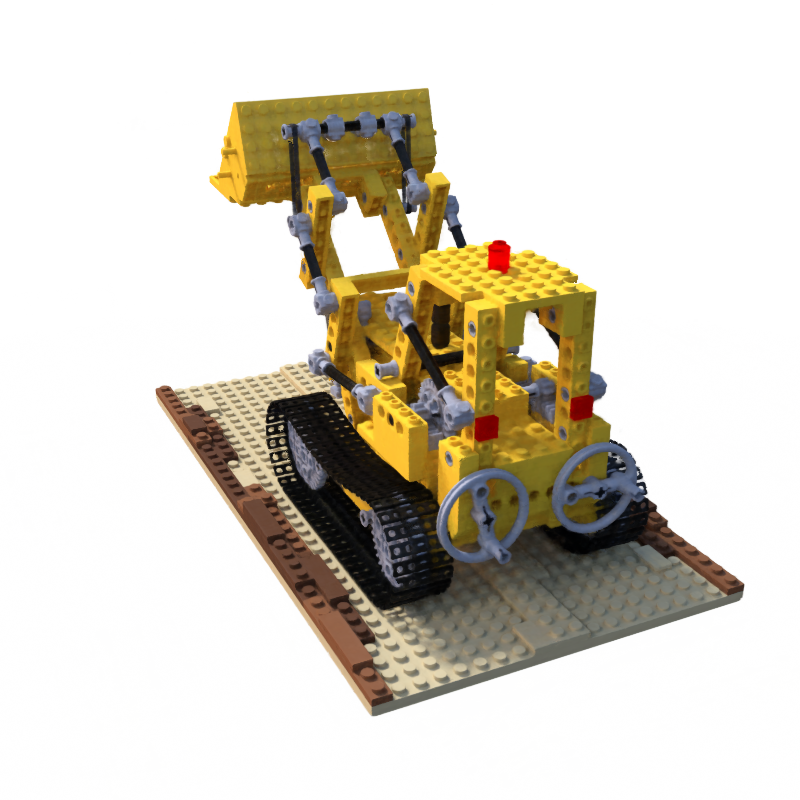} &
        \legopic{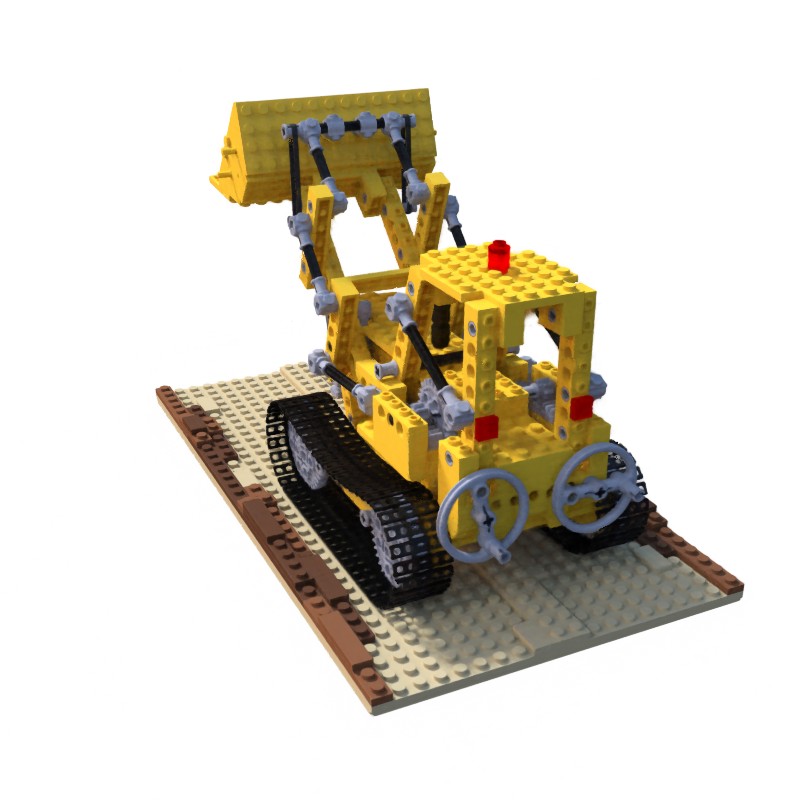}
    \\[-14mm]
    \rotatebox[origin=c]{90}{\hspace{18mm} Colors} &
        \legopic{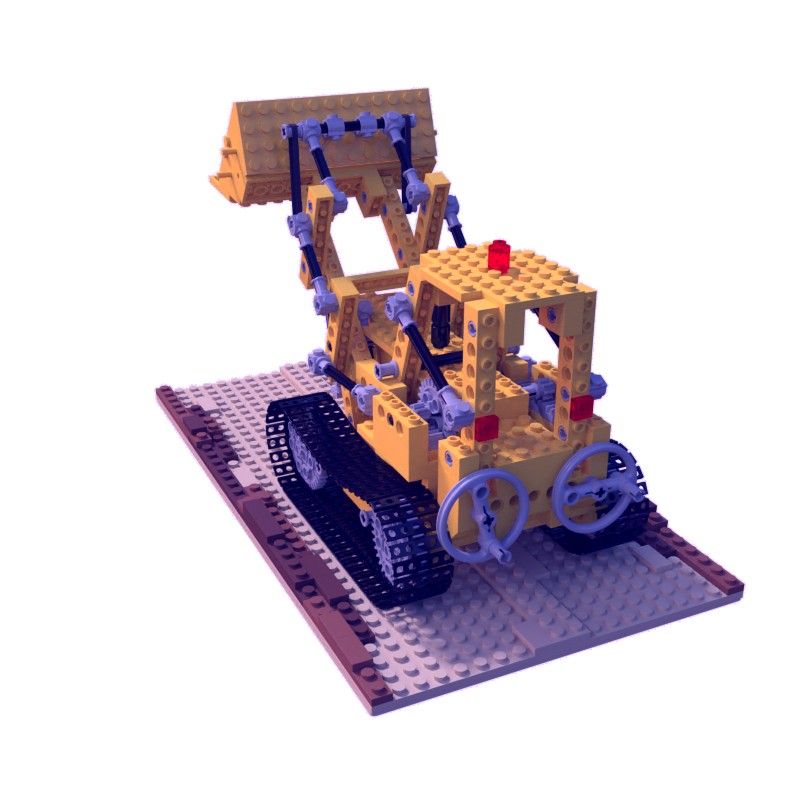} &
        \legopic{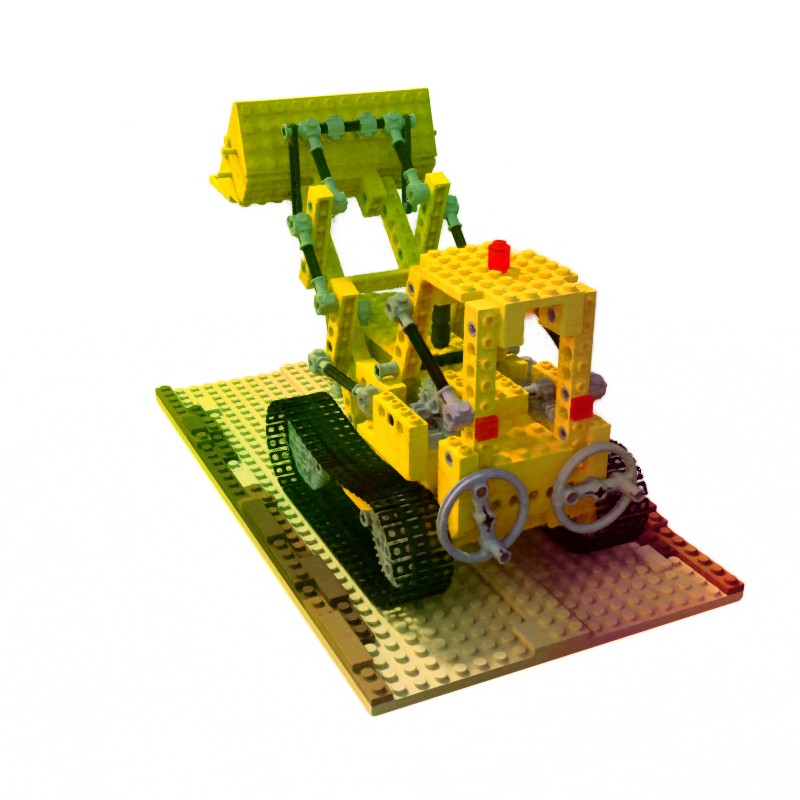} &
        \legopic{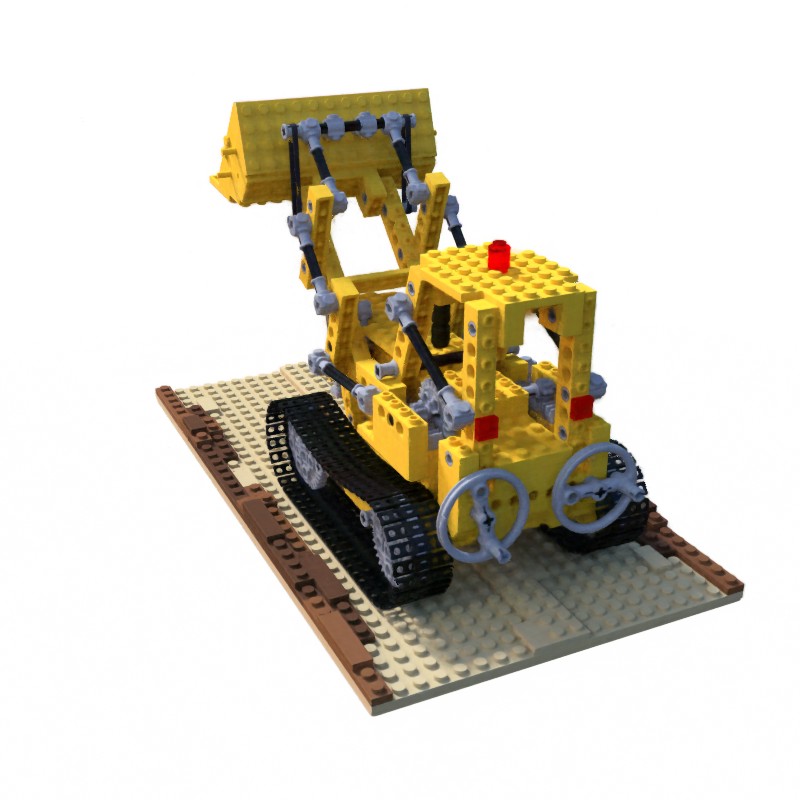} &
        \legopic{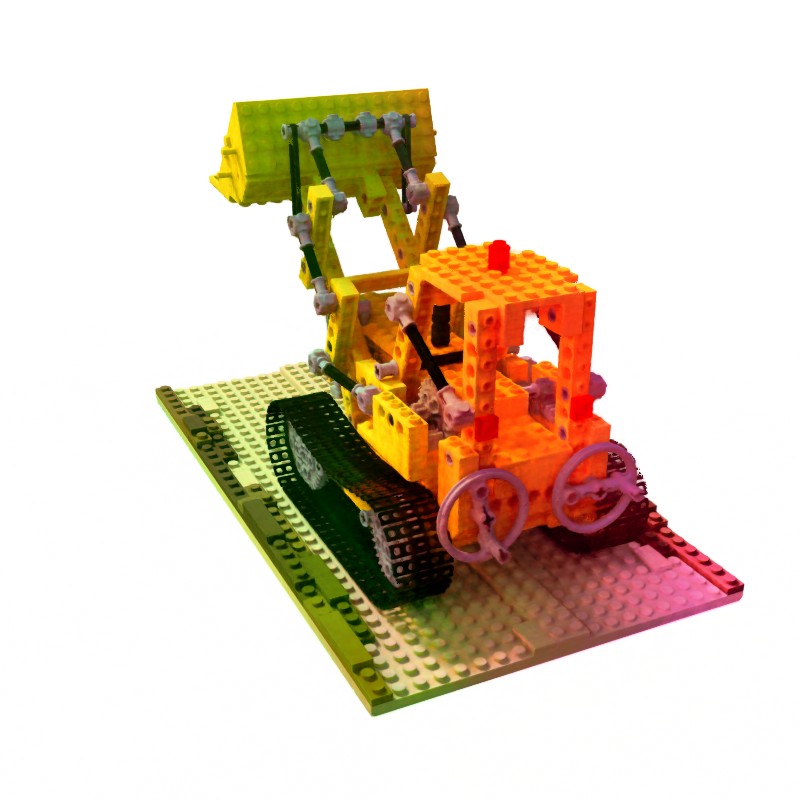} &
        \legopic{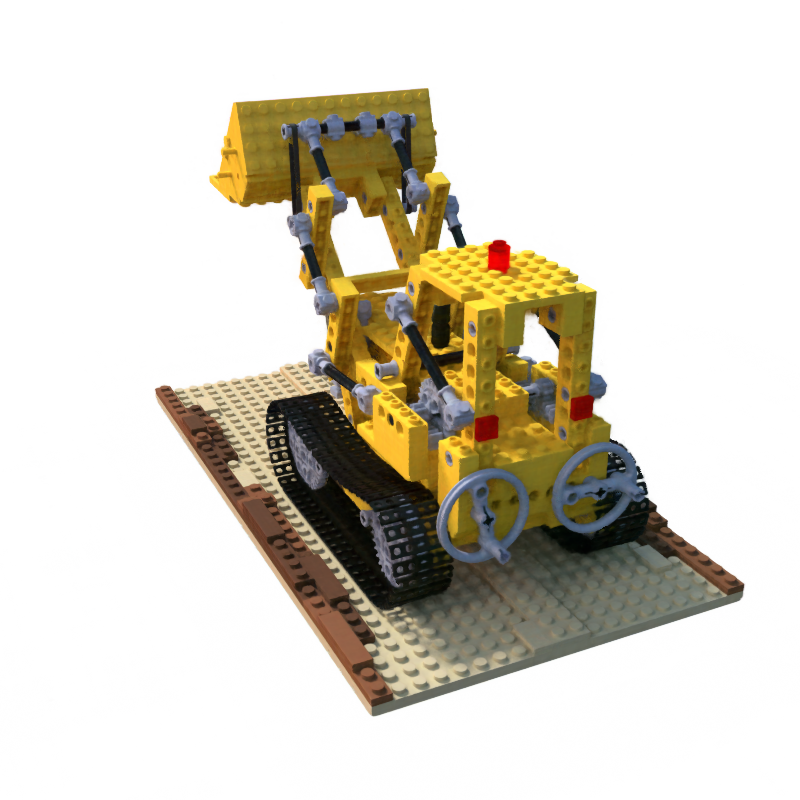}
    \\[-14mm]
    \rotatebox[origin=c]{90}{\hspace{18mm} Occluders} &
        \legopic{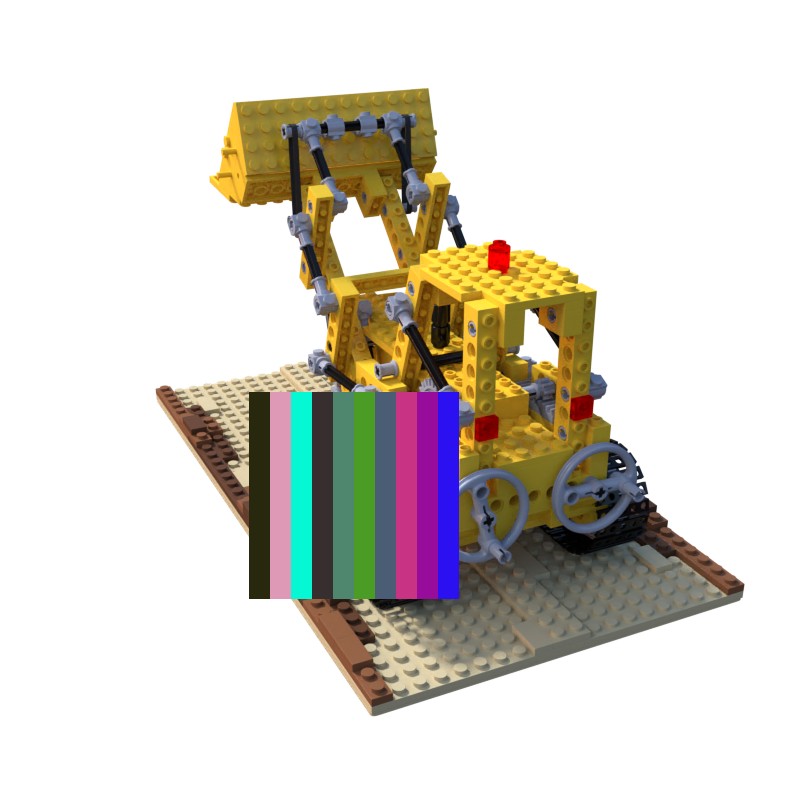} &
        \legopic{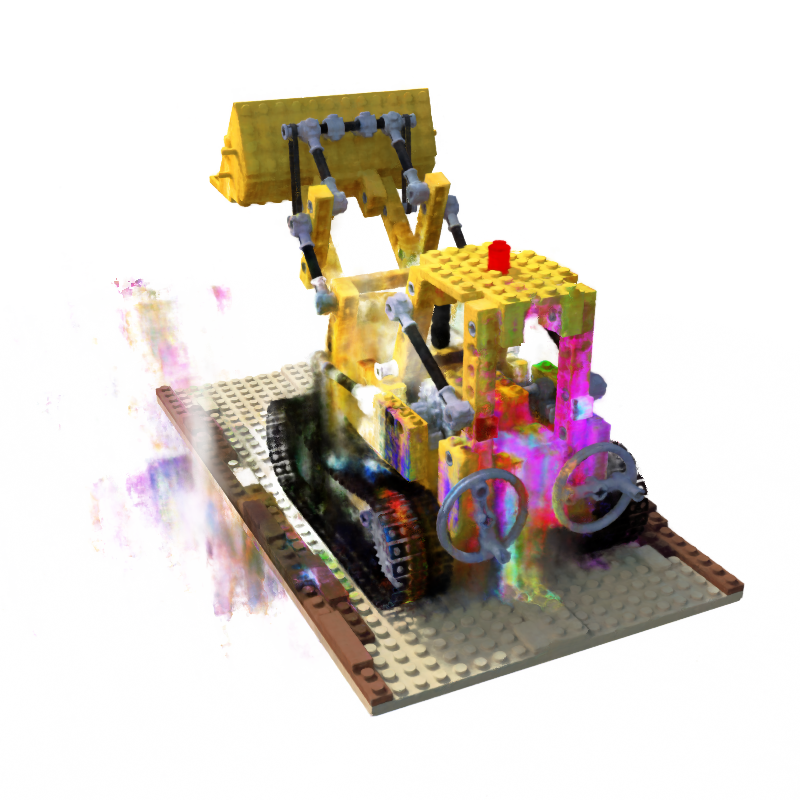} &
        \legopic{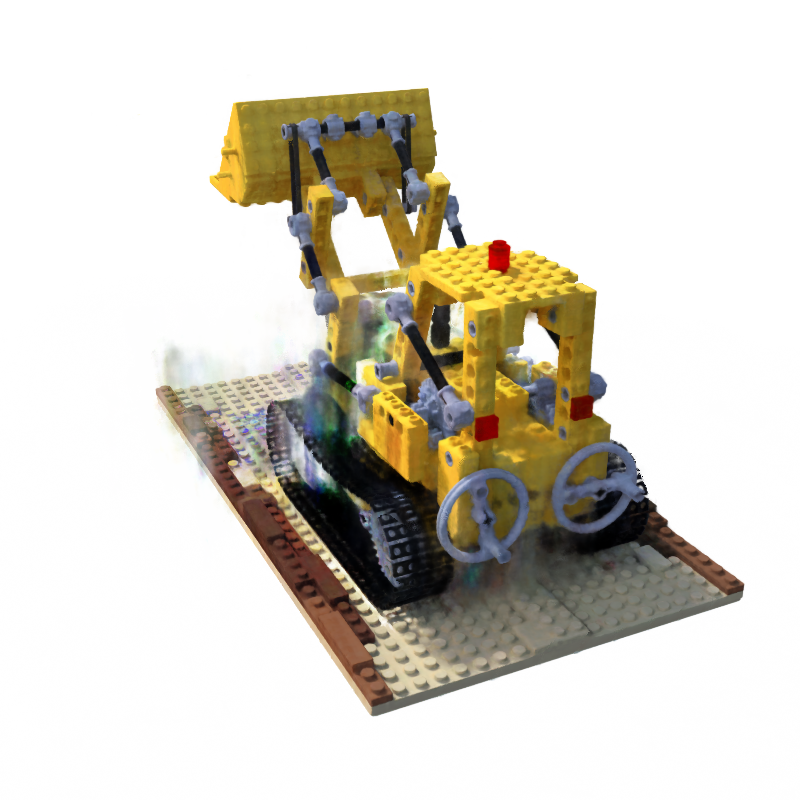} &
        \legopic{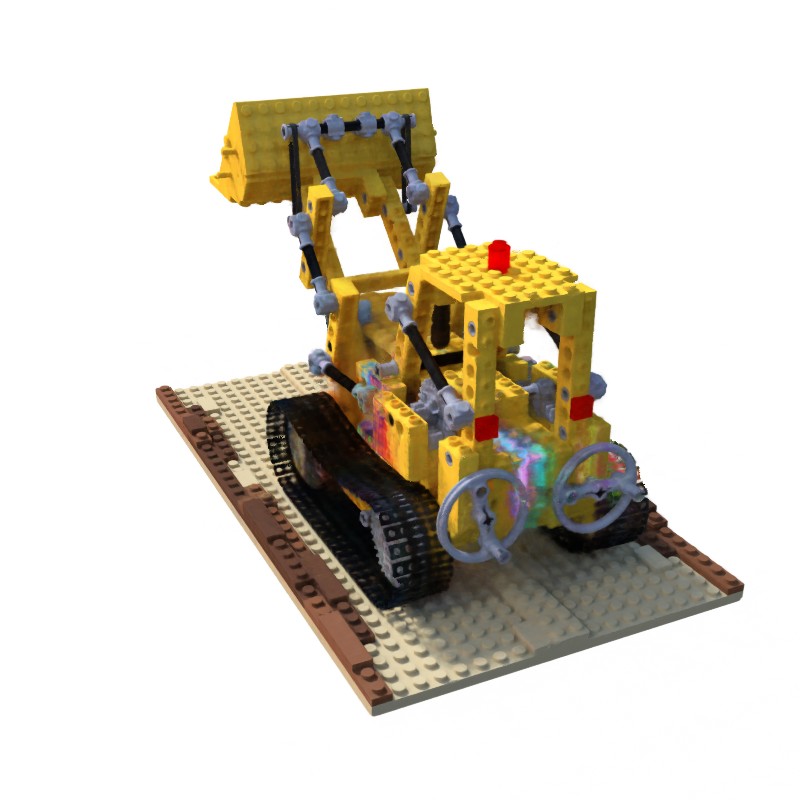} &
        \legopic{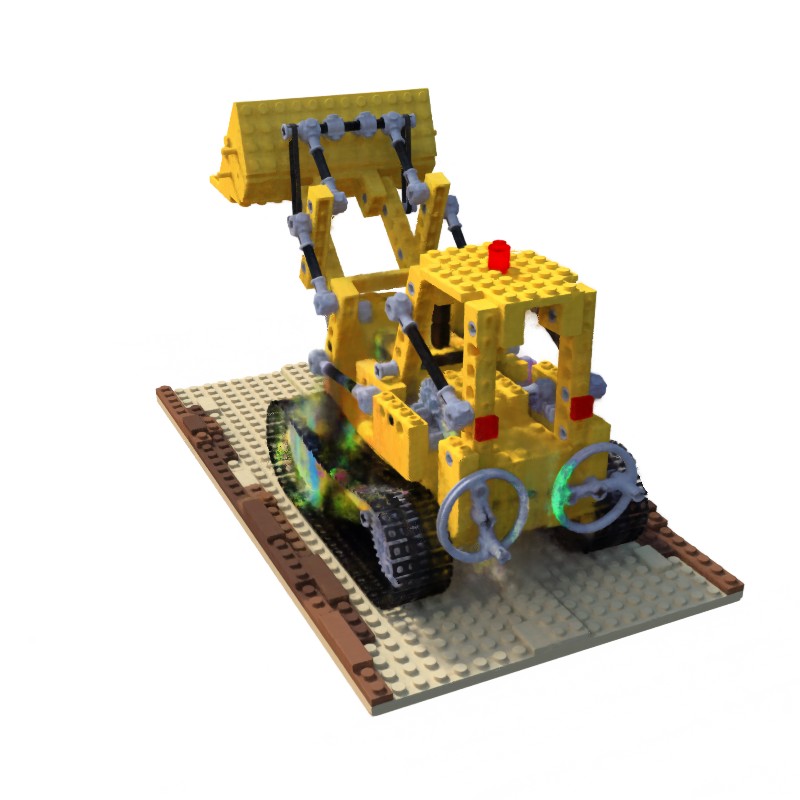}
    \\[-14mm]
    \rotatebox[origin=c]{90}{\hspace{18mm} Colors \& Occluders} &
        \legopic{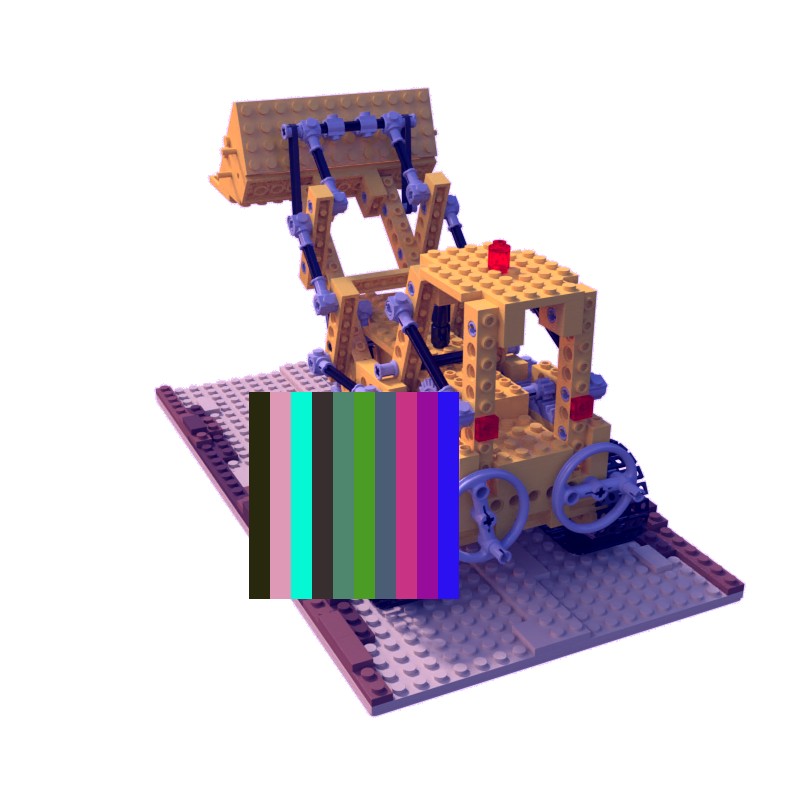} &
        \legopic{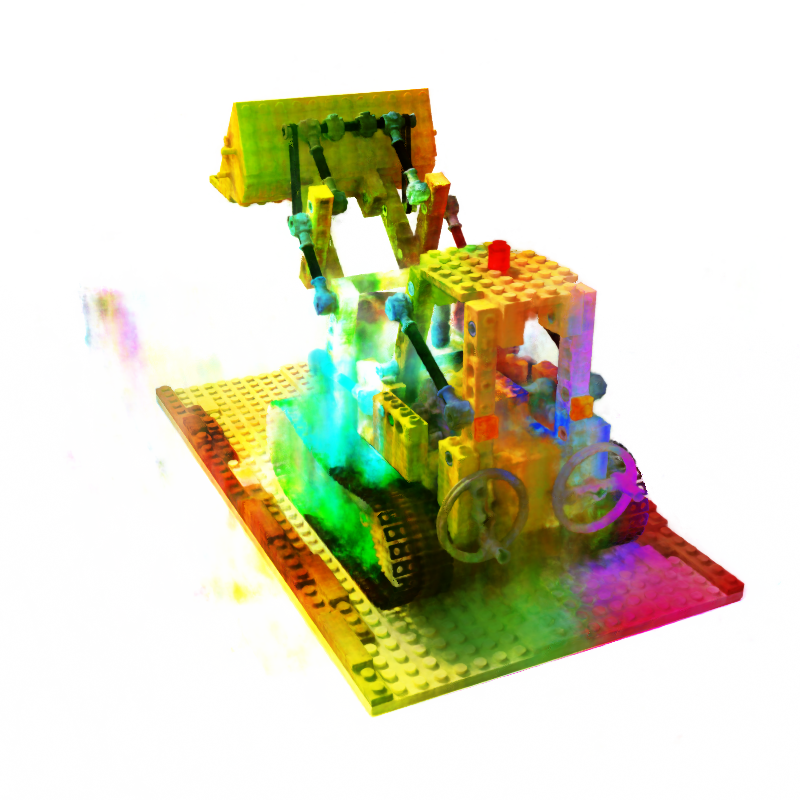} &
        \legopic{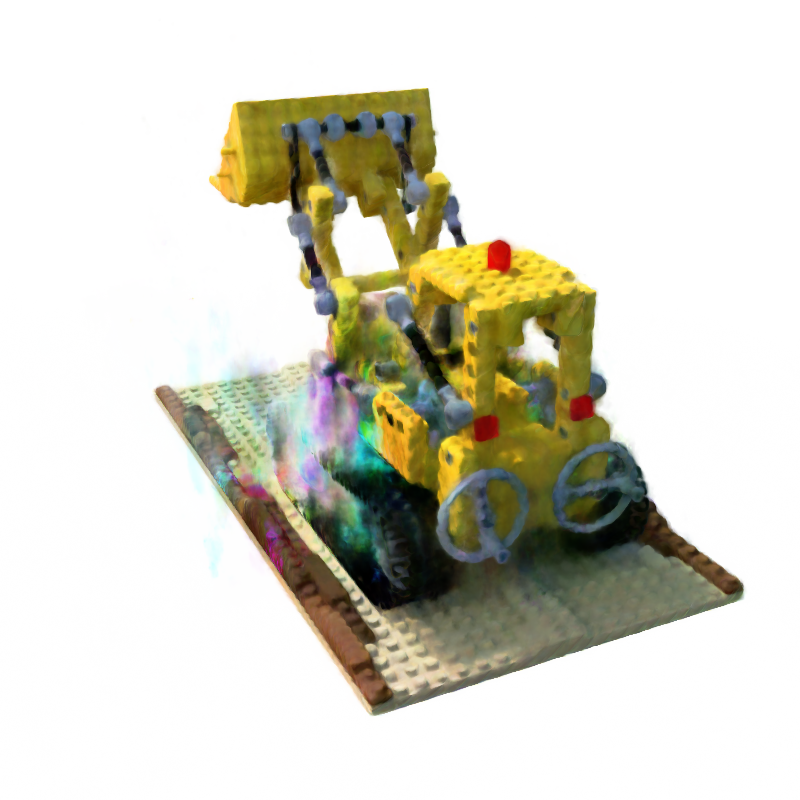} &
        \legopic{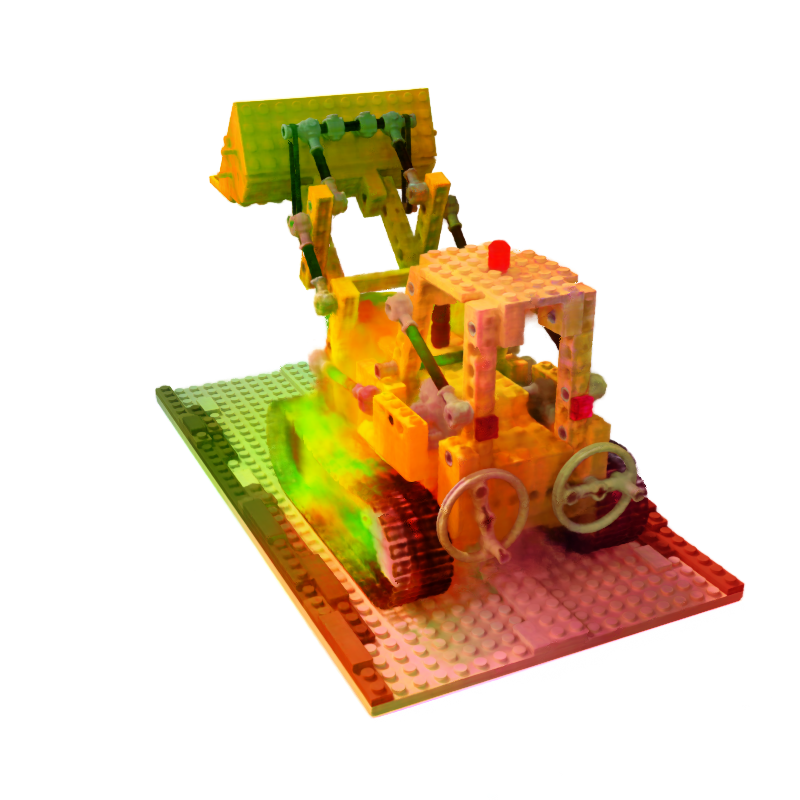} &
        \legopic{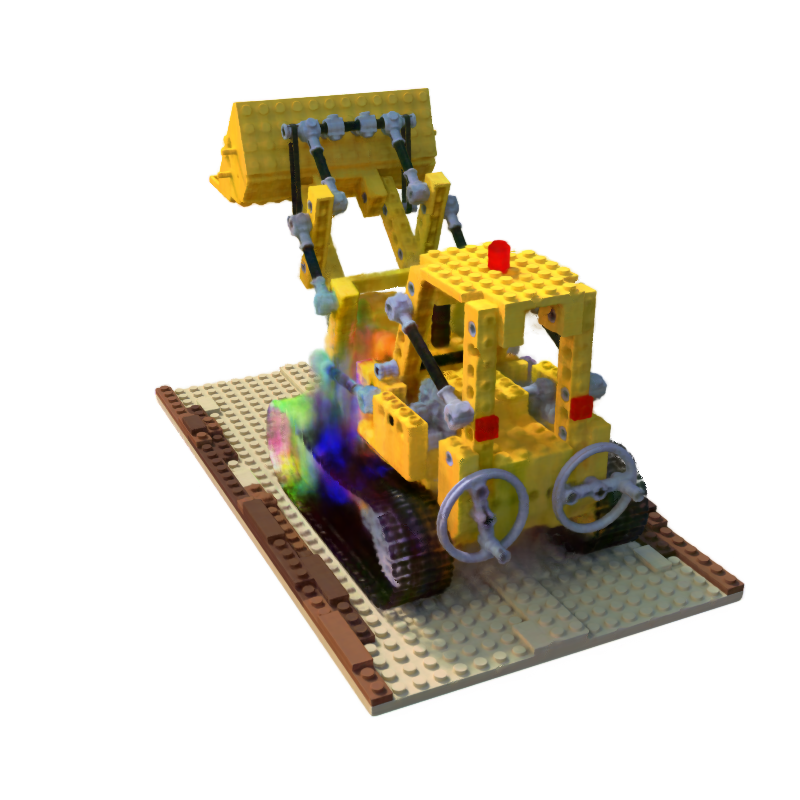}
    \\[-14mm]
    &
    Example perturbation &
    \multicolumn{1}{c}{\NERF{}} &
    \multicolumn{1}{c}{\NERFG{}} &
    \multicolumn{1}{c}{\NERFU{}} &
    \multicolumn{1}{c}{\NERFW{}}
    \end{tabular} 
    \caption{
        Example dataset perturbations and renderings from \NERF{}, \NERFG{}, \NERFU{} and \NERFW.
        The leftmost column illustrates the perturbations that were applied to the training dataset but using the test image for comparison.
        All other columns show renderings from models trained on datasets with each corresponding perturbation.
        \NERFG{} and \NERFU{} are largely able to disentangle color and occluder perturbations in isolation while \NERFW{} is able to do so simultaneously.
        \blendercc{Heinzelnisse}.
    }
    \label{fig:lego_qualitative}
\end{figure*}

\begin{figure*}
    \centering 
    \setlength{\tabcolsep}{-1.5pt}
    \def\arraystretch{1}
    \begin{tabular}{c@{\hskip 2mm}cccccc}
    \rotatebox[origin=c]{90}{\hspace{10mm} Brandenburg Gate} &
        \brandenburgpic{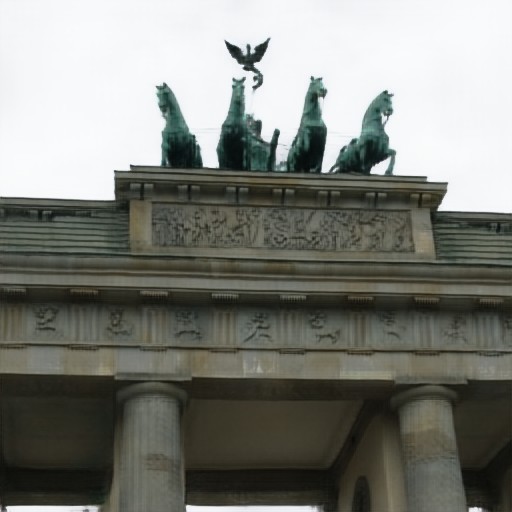} &
        \brandenburgpic{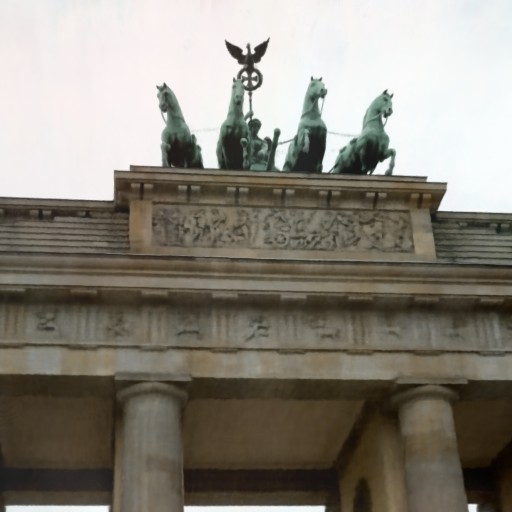} &
        \brandenburgpic{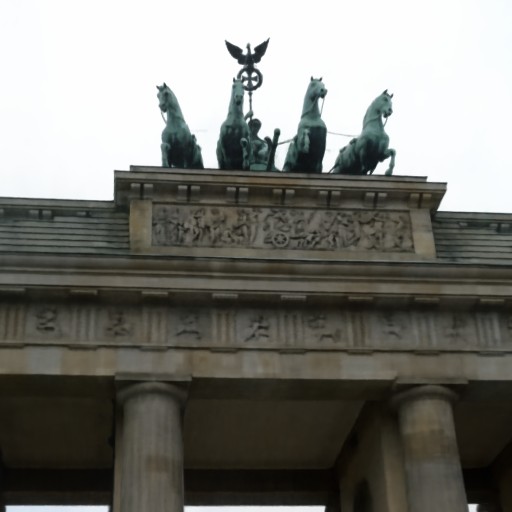} &
        \brandenburgpic{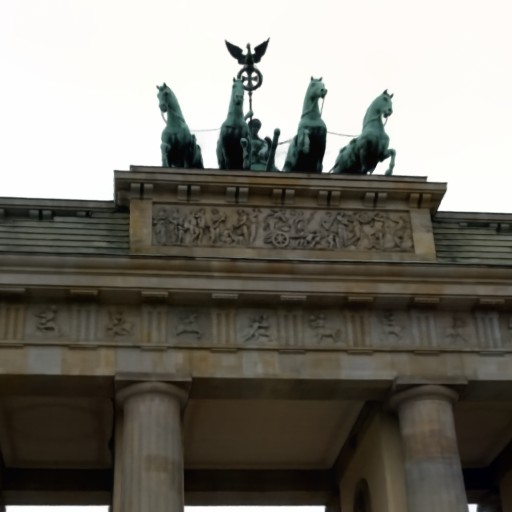} &
        \brandenburgpic{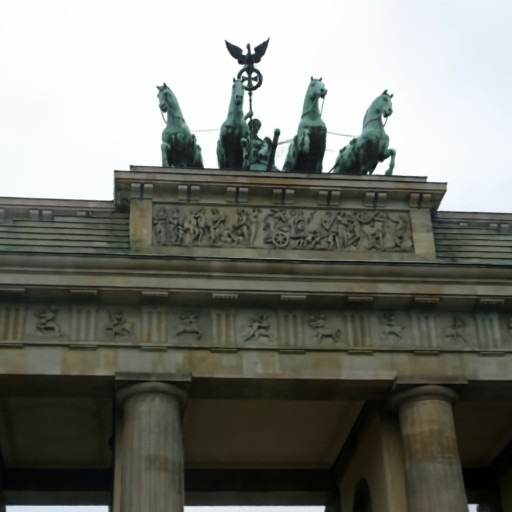} &
        \brandenburgpic{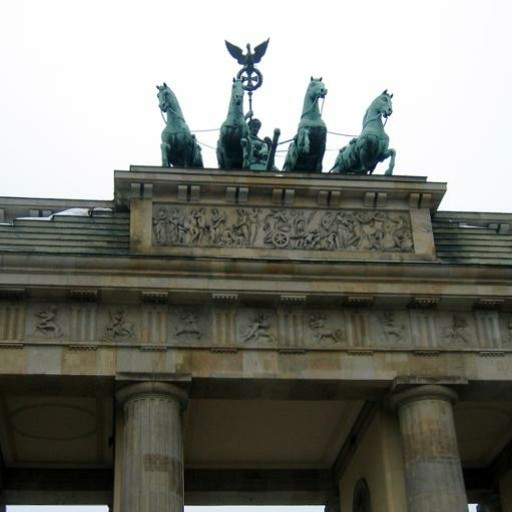}
    \\ [-11mm]
    \rotatebox[origin=c]{90}{\hspace{10mm} Trevi Fountain} &
        \trevipic{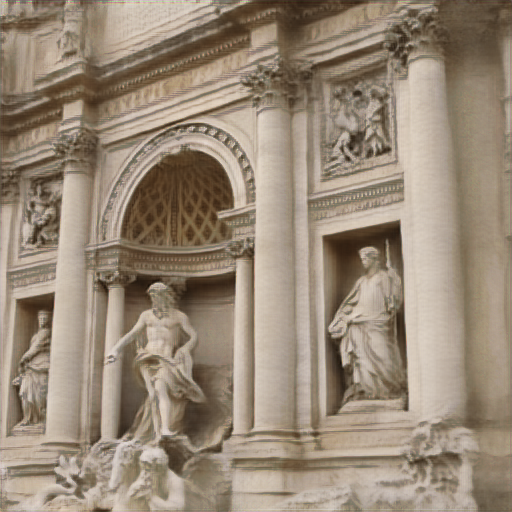} &
        \trevipic{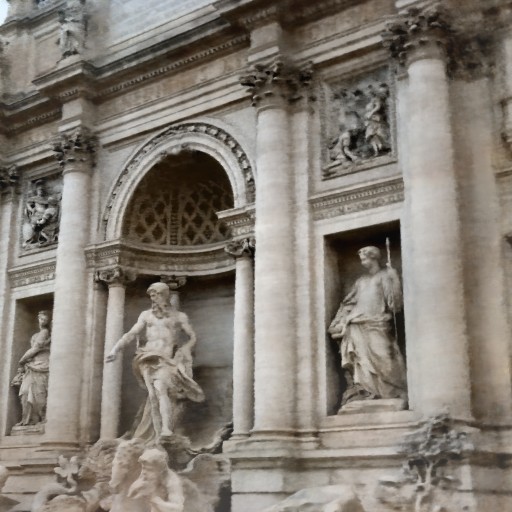} &
        \trevipic{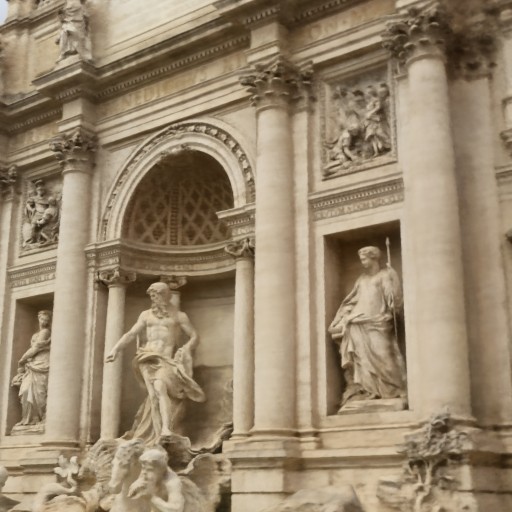} &
        \trevipic{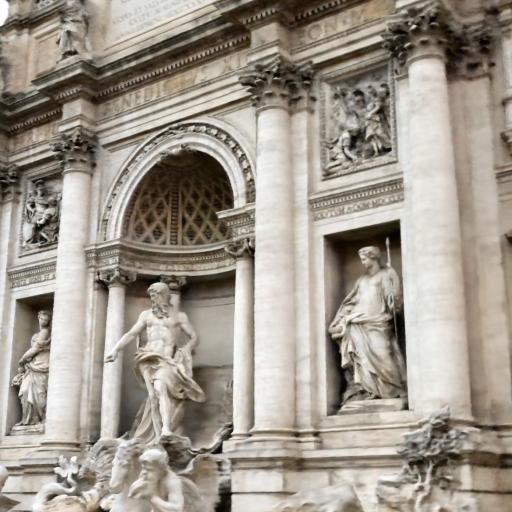} &
        \trevipic{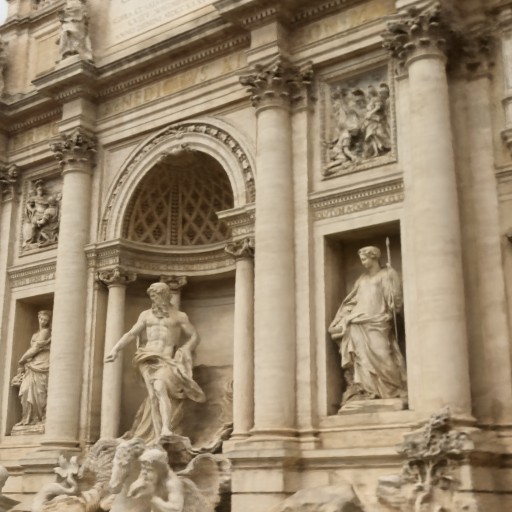} &
        \trevipic{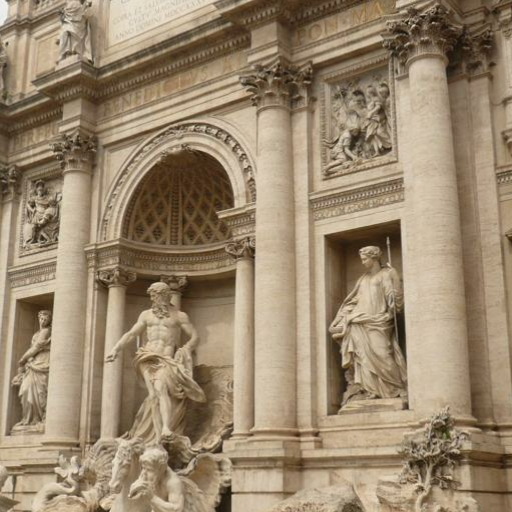}
    \\ [-11mm]
    \rotatebox[origin=c]{90}{\hspace{10mm} Hagia Sophia} &
        \hagiapic{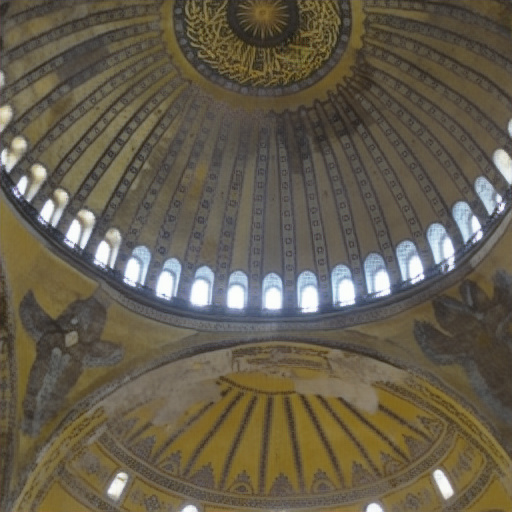} &
        \hagiapic{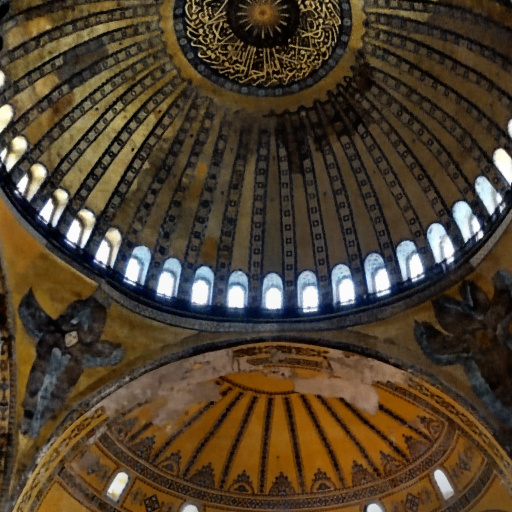} &
        \hagiapic{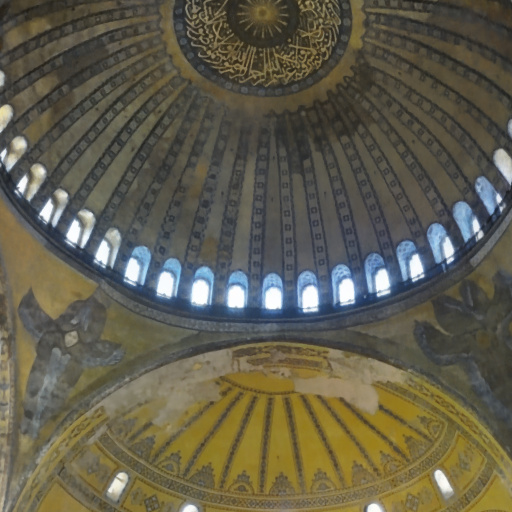} &
        \hagiapic{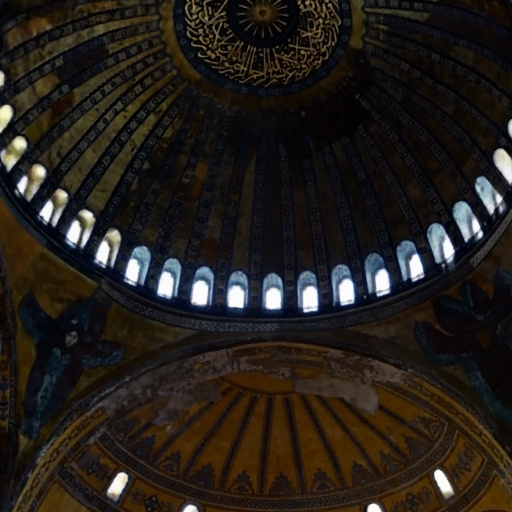} &
        \hagiapic{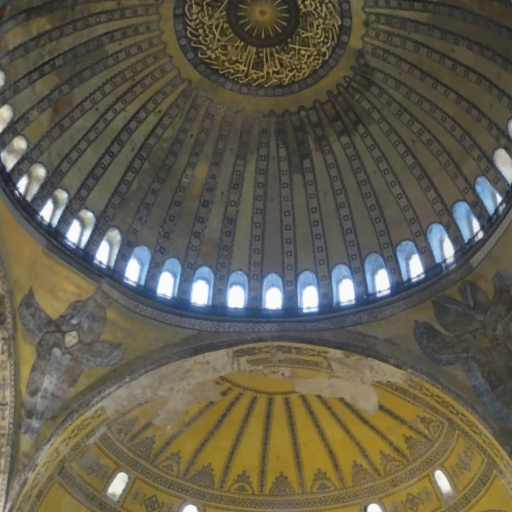} &
        \hagiapic{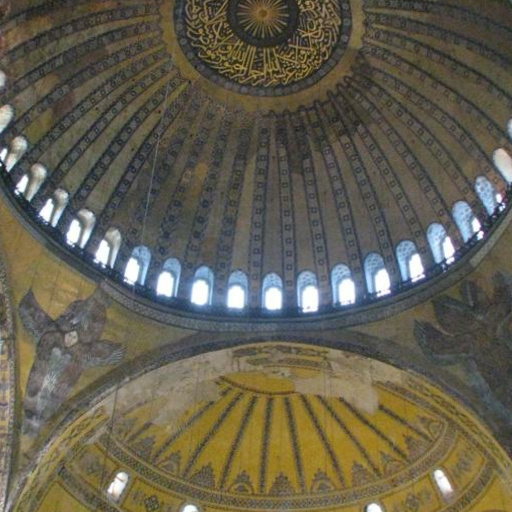}
    \\ [-11mm]
    &
    \multicolumn{1}{c}{NRW} &
    \multicolumn{1}{c}{\NERF{}} &
    \multicolumn{1}{c}{\NERFG{}} &
    \multicolumn{1}{c}{\NERFU{}} &
    \multicolumn{1}{c}{\NERFW{}} &
    \multicolumn{1}{c}{Ground-truth}
    \\
    \end{tabular} 
    \caption{
        Further qualitative results from experiments on Phototourism dataset.
        \NERFW{} is able to capture reflections (top row), consistent scene geometry at a distance (middle), and eliminate transient occluders (bottom).
        \flickrcc{yatani, jingjing, lricecsp}.
    } 
    \label{fig:phototourism_qualitative_2}
\end{figure*}

\end{document}